\documentclass[]{elsarticle}

\usepackage{lineno}
\usepackage{mathrsfs}
\usepackage{graphicx}
\usepackage{xcolor,colortbl}
\usepackage[algoruled,boxed,lined]{algorithm2e}
\usepackage{stmaryrd}
\usepackage{subcaption}
\usepackage{nicefrac}
\usepackage[cmex10]{amsmath}
\usepackage{amssymb}
\usepackage{algorithmic}
\usepackage{appendix}
\usepackage{hyperref}
\usepackage{ulem}
\usepackage{multirow}
\usepackage[margin=2.5cm]{geometry}

\renewcommand{\vec}[1]{\mathbf{#1}}

\newcommand*\diff{\mathop{}\!\mathrm{d}}

\newcommand{\norm}[1]{\left\lVert#1\right\rVert}

\newtheorem{thm}{Theorem}

\newtheorem{prop}{Proposition}

\newcommand{\E}{\mathbb{E}}

\newcommand{\F}{\mathcal{F}}
\newcommand{\X}{\mathcal{X}}

\newcommand{\D}{\mathcal{D}}
\newcommand{\A}{\mathcal{A}}
\newcommand{\Pk}{\mathcal{P}}

\newcommand{\x}{\vec x}

\newcommand{\y}{\vec y}
\newcommand{\w}{\vec w}

\newcommand{\tv}{\boldsymbol{t}}

\newcommand{\wrt}{w.\,r.\,t.~}
\newcommand{\eg}{e.\,g.~}

\journal{Information Sciences}

\begin{document}
	
	\begin{frontmatter}
		
		\title{Robust Unsupervised Domain Adaptation for Neural Networks via Moment Alignment}
		
		\author[jku]{Werner Zellinger\corref{cor1}}
		\ead{werner.zellinger@jku.at}
		\author[scch]{Bernhard A. Moser}
		\author[scch]{Thomas Grubinger}
		\author[jku]{Edwin Lughofer}
		\author[scch]{Thomas Natschl\"ager}
		\author[jku]{Susanne Saminger-Platz}
		
		\cortext[cor1]{Corresponding author}
		\address[jku]{Johannes Kepler University, Linz, Austria}
		\address[scch]{Software Competence Center Hagenberg GmbH, Hagenberg, Austria}
		
		\begin{abstract}
			A novel approach for unsupervised domain adaptation for neural networks is proposed.
			It relies on metric-based regularization of the learning process.
			The metric-based regularization aims at domain-invariant latent feature representations by means of maximizing the similarity between domain-specific activation distributions.
			The proposed metric results from modifying an integral probability metric such that it becomes less translation-sensitive on a polynomial function space.
			The metric has an intuitive interpretation in the dual space as the sum of differences of higher order central moments of the corresponding activation distributions.
			Under appropriate assumptions on the input distributions, error minimization is proven for the continuous case.
			As demonstrated by an analysis of standard benchmark experiments for sentiment analysis, object recognition and digit recognition, the outlined approach is robust regarding parameter changes and achieves higher classification accuracies than comparable approaches.\footnote{Source code available at \url{https://github.com/wzell/mann}.}
		\end{abstract}
		
		
	\end{frontmatter}
	
	\section{Introduction}
	\label{s:intro}
	{\it Transfer learning} focuses on solving machine learning problems by applying knowledge gained from different but related problems.
	In this work, the special case of {\it domain adaptation}~\cite{blitzer2006domain,pan2010survey,pan2011domain,ganin2016domain,li2017prediction} is considered.
	The goal of domain adaptation is to build a model that performs well on a \textit{target} data distribution while it is trained on a different but related \textit{source} data distribution.
	
	One important example is sentiment analysis of product reviews~\cite{glorot2011domain} where a model is trained on data of a source product category, \eg kitchen appliances, and it is tested on data of a related category, \eg books.
	A second example is the training of image classifiers on unlabeled real images by means of nearly-synthetic images that are fully labeled but have a different distribution~\cite{ganin2016domain}.
	Another example is the content-based depth range adaptation of unlabeled stereoscopic videos by means of labeled data from movies~\cite{zellinger2016linear}.
	
	A classifier's error on the target domain can be bounded in terms of its error on the source domain and a divergence between the source and the target domain distributions~\cite{ben2010theory}.
	This motivated many approaches to first extract features that overcome the distribution difference and subsequently minimize the source error~\cite{chen2012marginalized,pan2011domain}.
	Recently, approaches have been developed that embed domain adaptation in the feature learning process.
	One way to do this is to minimize a combined objective that ensures both a small source error and feature representations that overcome the domain difference~\cite{long2015learning,sun2016deep,ganin2016domain}.
	
	While much research has been devoted to the question of how to minimize the source error~\cite{lecun2015deep}, relatively little is known about objectives that ensure domain-invariant feature representations.
	In this contribution we focus on the latter question.
	In particular, we deal with the task of \textit{unsupervised domain adaptation} where no information is available about the target labels. However, the proposed approach is also applicable under the presence of target labels (\textit{semi-supervised domain adaptation}).
	
	We aim for a robust objective function, that is, (a) the convergence of our learning algorithm to sub-optimal solutions should guarantee similar domain-specific activation distributions and (b) the accuracy of our learning algorithm should be insensitive to changes of the hyper-parameters.
	The latter property is especially important in the unsupervised problem setting since the parameters must be selected without label information in the target domain and the application of parameter selection routines for hierarchical representation learning models can be computationally expensive.
	
	Our idea is to approach both properties by minimizing an integral probability metric~\cite{muller1997integral} between the domain-specific hidden activation distributions that is based on a polynomial function space of higher order.
	Although, the alignment of first and second order polynomial statistics performs well in domain adaptation~\mbox{\cite{tzeng2014deep,sun2016deep}} and generative modeling~\cite{mroueh2017mcgan}, higher order polynomials have not been considered before.
	One possible reason are instability issues that arise in the application of higher order polynomials.
	We solve these issues by modifying an integral probability metric such that it becomes translation-invariant on a polynomial function space.
	We call the metric the Central Moment Discrepancy (CMD).
	The CMD has an intuitive representation in the dual space as the sum of differences of higher order central moments of the corresponding distributions.
	We propose a robust domain adaptation algorithm for the training of neural networks that is based on the minimization of the CMD.
	The classification performance and accuracy sensitivity regarding parameter changes is analyzed on artificial data as well as on benchmark datasets for sentiment analysis of product reviews~\cite{chen2012marginalized}, object recognition~\cite{saenko2010adapting} and digit recognition~\cite{lecun1998mnist,netzer2011reading,ganin2016domain}.
	
	The main contributions of this work are as follows:
	\begin{itemize}
		\item We propose a novel approach for unsupervised domain adaptation for neural networks that is based on a metric-based regularization of the learning process. We call the metric the Central Moment Discrepancy (CMD).
		\item We prove several properties of the CMD including its computationally efficient implementable dual representation, a relation to weak convergence of distributions and a strictly decreasing upper bound for its moment terms.
		\item Our algorithm outperforms comparable approaches in standard benchmark experiments for sentiment analysis of product reviews, object recognition and digit recognition.
	\end{itemize}
	In addition, our approach is robust regarding the following aspects.
	\begin{itemize}
		\item Our approach overcomes instability issues of the learning process by solving the problem of mean over-penalization that arises in the application of integral probability metrics based on polynomial function spaces.
		\item In order to increase the visibility of the effects of the proposed method we refrain from hyper parameter tuning but carry out our experiments on $21$ domain adaption tasks with fixed regularization weighting parameter, fixed parameters of the metric, and without tuning of the learning rate.
		\item A post-hoc parameter sensitivity analysis shows that the classification accuracy of our approach is not sensitive to changes of the number-of-moments parameter and changes of the number of hidden nodes.
	\end{itemize}
	
	The paper is organized as follows:
	In Section~\ref{sec:related} we give a brief overview of related work.
	In Section~\ref{sec:da} we specify our model of domain adaptation and motivate the training of neural networks based on a joint objective that minimizes the source error and simultaneously enforces similar hidden activation distributions.
	Section~\ref{sec:prob_metric} presents the idea of applying the integral probability metric based on a polynomial function space and discusses the problem of mean over-penalization.
	In Section~\ref{sec:cmd} we propose the CMD and in Section~\ref{sec:properties} we analyze some convergence properties.
	A gradient based algorithm for domain adaptation that minimizes the CMD is presented in Section~\ref{sec:MANN}.
	Section~\ref{sec:experiments} analyzes the classification performance and the parameter sensitivity of our algorithm based on benchmark datasets.
	Section~\ref{sec:conclusion} concludes the work.
	
	\section{Related Work}
	\label{sec:related}
	
	Many methods have been proposed for approaching the problem of domain adaptation.
	Some emphasize the analysis of linear hypotheses~\cite{blitzer2006domain,baktashmotlagh2013unsupervised,cortes2014domain,nikzad2018nipals} whereas more recently non-linear representations have been studied~\cite{glorot2011domain}, including neural networks~\cite{long2015learning,ganin2016domain,sun2016deep,long2016unsupervised,long2016joint,bousmalis2016domain,tzeng2017adversarial}.
	In the latter case, the source and the target domain distributions are aligned in the latent activation space in order to guarantee domain-invariant feature representations.
	Three prominent research directions can be identified for the choice of the alignment objective.
	
	The first research direction investigates the re-weighting of the neural network activations such that specific mean and covariance features are aligned. These approaches work particularly well in the area of object recognition~\cite{li2016revisiting} and text classification~\cite{sun2016return}.
	Mean and covariance feature alignment has been extended to the minimization of the Frobenius norm between the covariance matrices of the neural network activations~\cite{sun2016deep}.
	This distance function is parameter-free and it does not require additional unsupervised validation procedures neither parameter heuristics.
	We show that these approaches can be further improved in terms of time complexity and prediction accuracy by additionally considering moment characteristics of higher orders.
	
	Another research direction investigates the minimization of the \textit{Proxy-$\mathcal{A}$ distance}~\cite{ben2010theory} for distribution alignment.
	This distance function is theoretically motivated and can be implemented by means of an additional classifier with the objective of separating the distributions. For distribution alignment, the gradient of the classifier is reversed during back-propagation~\cite{ganin2016domain,tzeng2017adversarial}.
	Unfortunately, an additional classifier must be trained in this approach, which includes the need for new parameters, additional computation times and validation procedures.
	In addition, the reversal of the gradient causes several theoretical problems~\cite{arjovsky2017towards} that contribute to instability and saturation during training.
	Our approach achieves higher classification accuracy on several domain adaptation tasks in benchmark datasets.
	
	A third research direction applies a distance function called Maximum Mean Discrepancy (MMD)~\cite{gretton2006kernel}.
	It is an integral probability metric that is based on the unit ball of a reproducing kernel Hilbert space (RKHS).
	Different underlying kernel functions lead to different RKHSs and therefore to different versions of the MMD.
	There exist approaches that are based on linear kernels~\cite{tzeng2014deep} that can be interpreted as mean feature matching.
	A combination of Gaussian kernels is used~\cite{long2015learning} to tackle the sensitivity of the MMD \wrt changes of the Gaussian kernel parameter. This is done by means of a combination of different kernels with heuristically selected parameters.
	In addition, the approach comes with the theoretical knowledge from the studies about RKHSs~\cite{fukumizu2009kernel} and a linear-time implementation.
	We solve the problem of the high sensitivity of the MMD \wrt the kernel parameter by an alternative distance function that is less sensitive to changes of its parameter.
	
	Some recent approaches focus on combining research about specific neural network architectures with the application of the MMD using Gaussian kernel~\cite{bousmalis2016domain,long2016unsupervised,long2016joint}.
	Our approach is not restricted to multiple layers or network architectures.
	Actually, it can be combined with these ideas.

	\section{Problem Description of Domain Adaptation}
	\label{sec:da}
	Without loss of generality, let us formulate the problem of unsupervised domain adaptation for binary classification~\cite{ben2010theory,ganin2016domain,mansour2009domain}.
	We define a \textit{domain} as a pair $\left\langle \mathcal{D}, g\right\rangle$ of a distribution $\cal D$ on the input space $\cal X$ and a labeling function $g:{\cal X}\rightarrow [0,1]$, which can have intermediate values when labeling occurs non-deterministically. We denote by $\left\langle {\cal D}_S,g_S \right\rangle $ the \textit{source} domain and by $\left\langle {\cal D}_T,g_T \right\rangle $ the \textit{target} domain.
	In order to measure to what extent a classifier $h:{\cal X}\rightarrow [0,1]$ disagrees with a given labeling function $g$, we consider the expectation of its difference \wrt the distribution ${\D_A}$,
	\begin{equation}
		\label{eq:error}
		\epsilon_A(h,g)=\mathbb{E}_{\D_A}\big[|h-g|\big]
	\end{equation}
	where $\E_{\mathcal{D}_A}[f]=\int_\mathcal{X} f(\x) d \mathcal{D}_A$.
	We refer to $\epsilon_S(h,g_S)$ as the source error and to $\epsilon_T(h,g_T)$ as the target error.
	
	In our problem setting, two samples are given: a labeled source sample $S=\{(\x_i, g_S(\x_i))\}_{i=1}^m\subseteq\X\times[0,1]$ with ${\x_i\sim\D_S}$ and an unlabeled target sample $T=\{\x_j\}_{j=1}^n\subseteq\X$ with $\x_j\sim\D_T$.
	The goal of unsupervised domain adaptation is to build a classifier $h:\X\rightarrow [0,1]$ with a low target error $\epsilon_T(h, g_T)$ while no information about labels in the target domain is given.
	
	\subsection{Motivation for Unsupervised Domain Adaptation}
	\label{sec:target_error}
	To motivate exploration of the problem of unsupervised domain adaptation, let us first show how the error minimization in the target domain relates to the minimization of the source error and the difference between the domains $\left\langle {\cal D}_S,g_S \right\rangle $ and $\left\langle {\cal D}_T,g_T \right\rangle $.
	
	In practice, we expect the dissimilarity between the labeling functions $g_S$ and $g_T$ to be small~\cite{ben2010theory} or even zero~\cite{sugiyama2008direct}.
	Otherwise, there is no way to infer a good estimator based on the training sample.
	Therefore, we focus on the distance between the distributions $\D_S$ and $\D_T$.
	A suitable class of distance measures consists of integral probability metrics~\cite{muller1997integral}.
	Given a function class $\F=\{f:\X\rightarrow\mathbb{R}\}$, an integral probability metric is defined by
	\begin{equation}
		\label{eq:ipm}
		d_\F(\D_S,\D_T)=\sup_{f\in\F}\big|\E_{\D_S}[f] - \E_{\D_T} [f]\big|
	\end{equation}
	with $\E_{\D_S}[f]$ as in Eq.\eqref{eq:error} and the supremum $\sup$.
	Integral probability metrics play an important role in probability theory~\cite{zolotarev1983probability} and statistics~\cite{gretton2006kernel}.
	An integral probability metric is a pseudo-metric and it is a metric if and only if the function class $\F$ separates the set of all signed measures $\mu$ with $\mu(\X)=0$~\cite[page 432]{muller1997integral}.
	
	Based on these results a bound on the classifier's target error may be determined. Following the proof of~\cite[Theorem~1]{ben2010theory}, we may state the following result:
	\begin{thm}[Ben-David et al., 2006]
		\label{thm:targeterror}
		Let $h:\X\rightarrow [0,1]$ be a classifier, then
		\begin{align}
			\label{eq:bound}
			\epsilon_T(h, g_T) \leq~ &\epsilon_S(h, g_S) + d_\F(\D_S,\D_T) \\
			&+ \min\left\lbrace \E_{\D_S}\big[|g_S-g_T|\big],
			\E_{\D_T}\big[|g_S-g_T|\big]\right\rbrace \nonumber
		\end{align}
		with a suitable function class $\F$ that contains $\left|h-g_S\right|$ and $\left| h-g_T \right|$.
	\end{thm}
	
	Under the assumption on the difference between $g_S$ and $g_T$ to be small, Theorem~\ref{thm:targeterror} shows that the source error is a good indicator for the target error if the two domain distributions $\D_S$ and $\D_T$ are similar with respect to an integral probability metric defined in Eq.\eqref{eq:ipm}.
	
	\subsection{Domain Adaptation with Neural Networks}
	\label{sec:da_w_distr_al}
	As an example let us consider a continuous model of a neural network classifier $h=h_1\circ h_0$ consisting of a representation learning part $h_0:\X\rightarrow\A$ from the inputs $\X\subset\mathbb{R}^m$ to the activations $\A\subset\mathbb{R}^m$, e.g. a deep neural network, and a classification part $h_1:\A\rightarrow[0,1]$ from the activations to the labels $[0,1]$. Assume $h_0$ to be an invertible function from $\X$ to $\A$, e.g.~\cite{jacobsen2018revnet}.
	Then, for each continuous function $p$ and distribution $\D$, the ``change of variables'' theorem~\cite[Theorem 4.1.11]{dudley2002real} yields
	\begin{equation}
		\int_{\X} p\circ h_0\diff \D = \int_\A p\diff (h_0\circ \D),
	\end{equation}
	which implies that
	\begin{equation}
		\label{eq:h0metric}
		d_\F(\D_S,\D_T) = d_\mathcal{P}(h_0\circ \D_S,h_0\circ \D_T),
	\end{equation}
	where $\mathcal{P}=\{h_0\circ f| f\in\F\}$. 
	
	Eq.\eqref{eq:h0metric} allows us to minimize $\epsilon_T(h, g_T)$ in Eq.\eqref{eq:bound} by aligning the distribution of the activations $h_0\circ\D_S$ and $h_0\circ\D_T$ rather than the domain distributions $\D_S$ and $\D_T$.
	In addition, Eq.\eqref{eq:h0metric} allows us to focus on simple function classes $\mathcal{P}$, \eg polynomials, that are suitable for the alignment of neural networks activation distributions $h_0\circ\D_S$ and $h_0\circ\D_T$ rather than considering complex function classes $\mathcal{F}$ that are suitable for general domain distributions $\D_S$ and $\D_T$.
	
\begin{figure}[h]
		\centering
		\includegraphics[width=\textwidth]{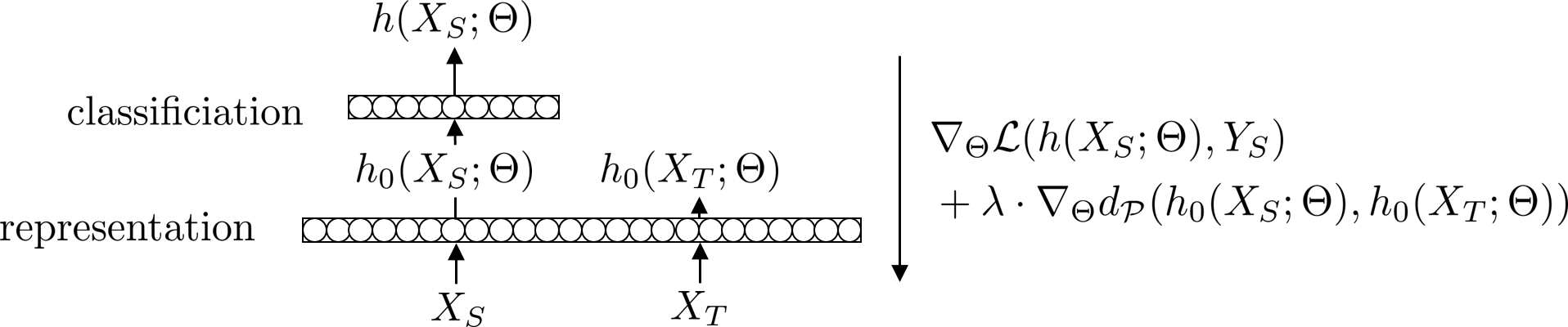}
		\caption[Architecture]{Schematic sketch of a feed-forward neural network $h(X;\Theta)$ with parameters $\Theta$ optimized via gradient descent based on the minimization of a source loss $\mathcal{L}(h(X_S;\Theta),Y_S)$ and the minimization of a distance $d_\mathcal{P}$ between the activations $h_0(X_S;\Theta)$ and $h_0(X_T;\Theta)$ of the source sample $X_S$ and the target sample $X_T$, where 
			$Y_S$ denotes the labels in the source domain. The minimization of $d_\mathcal{P}$ ensures domain-invariant representations.
			$\nabla_\Theta$ refers to the gradient \wrt $\Theta$ and $\lambda$ denotes the domain regularization parameter.}
		\label{fig:architecture}
	\end{figure}

	A realization of this idea based on gradient descent is shown in Fig.~\ref{fig:architecture}.
	
	\section{Integral Probability Metric on a Polynomial Function Space}
	\label{sec:prob_metric}
	
	Depending on the choice of the function set $\F$ for the integral probability metrics in Eq.\eqref{eq:ipm} one might obtain the Wasserstein distance, the total variation distance, or the Kolmogorov distance.
	In our approach, we focus on polynomial function spaces.
	The expectations of polynomials are sums of moments.
	This allows a natural interpretation of how the function set $\mathcal{F}$ in Eq.\eqref{eq:ipm} acts on the activation distributions.
	In applications such as image retrieval, moments are known as robust distribution descriptors~\cite{rui1999image}.
	
	Let us consider the vector-valued function
	\begin{equation}
		\label{eq:nu}
		\begin{alignedat}{2}
			\boldsymbol{\nu}^{(k)}:\mathbb{R}^m&\longrightarrow &&\mathbb{R}^{\frac{(k+1)^{m-1}}{(m-1)!}} \\
			\vec x&\longmapsto &&\Big( x_1^{r_1}\cdots x_m^{r_m} \Big)_{\substack{(r_1, \ldots, r_m)\in\mathbb{N}_0^m\\r_1+\ldots+r_m=k}}
		\end{alignedat}
	\end{equation}
	mapping a $m$-dimensional vector $\x=\big(x_1,\ldots,x_m\big)$ to its $\frac{(k+1)^{m-1}}{(m-1)!}$ monomial values $x_1^{r_1}\cdots x_m^{r_m}$ of order $k=r_1+\ldots+r_m$ with ${(r_1,\ldots,r_m)\in\mathbb{N}^m_0}$ using the lexicographic order for its elements, e.g. $\boldsymbol{\nu}^{(3)}((x_1, x_2))= (x_1^3, x_1^2 x_2, x_1 x_2^2, x_2^3)$.
	
	Further, let us denote by $\mathcal{P}^k$ the class of homogeneous polynomials ${p:\mathbb{R}^m\rightarrow\mathbb{R}}$ of degree $k$ with normalized coefficient vector, i.e.
	\begin{align}
		\label{eq:polyspace}
		p(\vec x) =&~ \langle \vec{w}, \boldsymbol{\nu}^{(k)}(\vec x)\rangle_2,
	\end{align}
	with $\| \vec{w}\|_2 \leq 1$ for the real vector $\vec{w}$ and the euclidean norm $\|.\|_2$.
	For example, the expectations of polynomials in $\mathcal{P}^3$ \wrt a distribution $\D$ are corresponding linear combinations of the third raw moments of $\D$, i.e.
	\begin{align}
		\E[p(\vec x)]= w_1 \E_\D[x_1^3]+ w_2 \E_\D[x_1^2 x_2]+ w_3 \E_\D[x_1 x_2^2]+ w_4 \E_\D[x_2^3],
	\end{align}
	with $\sqrt{w_1^2+w_2^2+w_3^2+w_4^2}\leq 1$.
	
	{It is interesting to point out that the space of polynomials $\mathcal{P}^k$ in Eq.\eqref{eq:polyspace} is the unit ball of a reproducing kernel Hilbert space.}

	\subsection{The Problem of Mean Over-Penalization}
	\label{subsec:mean_problem}
	
	{Unfortunately, an integral probability metric in Eq.\eqref{eq:ipm} based on the function space $\Pk^k$ in Eq.\eqref{eq:polyspace} and different other metrics~\mbox{\cite{mroueh2017mcgan,li2017mmd}} suffer from the drawback of mean over-penalization which becomes worse with increasing polynomial order.}
	For the sake of illustration, let us consider two distributions $\D$ and $\D'$ on $\mathbb{R}$.
	For $k=1$ we obtain
	\begin{align}
		\label{eq:muDiff}
		d_{\mathcal{P}^1}(\D,\D')  =~ &\sup_{|\omega|\leq 1} \big|\E_{\D}[\omega\, x]  - \E_{\D'}[\omega\,x]\big| \nonumber \\
		=~ &|\mu - \mu'|,
	\end{align}
	where $\mu = \E_{\D}[x]$ and $\mu' = \E_{\D'}[x]$. 
	Now, let us consider higher orders $k \in \mathbb{N}$.
	Assume that the distributions $\D$ and $\D'$ have identical central moments
	$c_j(\D) := \E[(x - \mu)^j]$ but different means $\mu  \neq \mu'$.
	By expressing the raw moment $\E_{\D}[x^k]$ by its central moments 
	$c_j(\D)$, we obtain, by means of the binomial theorem,
	\begin{align}
		\label{eq:meansensitive}
		d_{\mathcal{P}^k}(\D,\D') & =  \big|\E_\D[x^k]-\E_{\D'}[x^k]\big| \nonumber\\
		& = \left|\sum_{j=0}^k\binom{k}{j} c_j(\D) (\mu^{k-j} - {\mu'}^{k-j})\right|.
	\end{align}
	Since the mean values contribute to the sum of Eq.\eqref{eq:meansensitive} by its powers, the metric in Eq.\eqref{eq:ipm} with polynomials as function set is not translational invariant.
	Much worse, consider for example $\mu = 1 +\varepsilon/2$ and $\mu' = 1 -\varepsilon/2$, 
	then small changes of the mean values can lead to large deviations in the resulting metric, i.e. causing instability in the learning process.
	
\begin{figure}
		\makebox[\linewidth][c]{%
			\begin{subfigure}[b]{.40\textwidth}
				\centering
				\includegraphics[width=.95\textwidth]{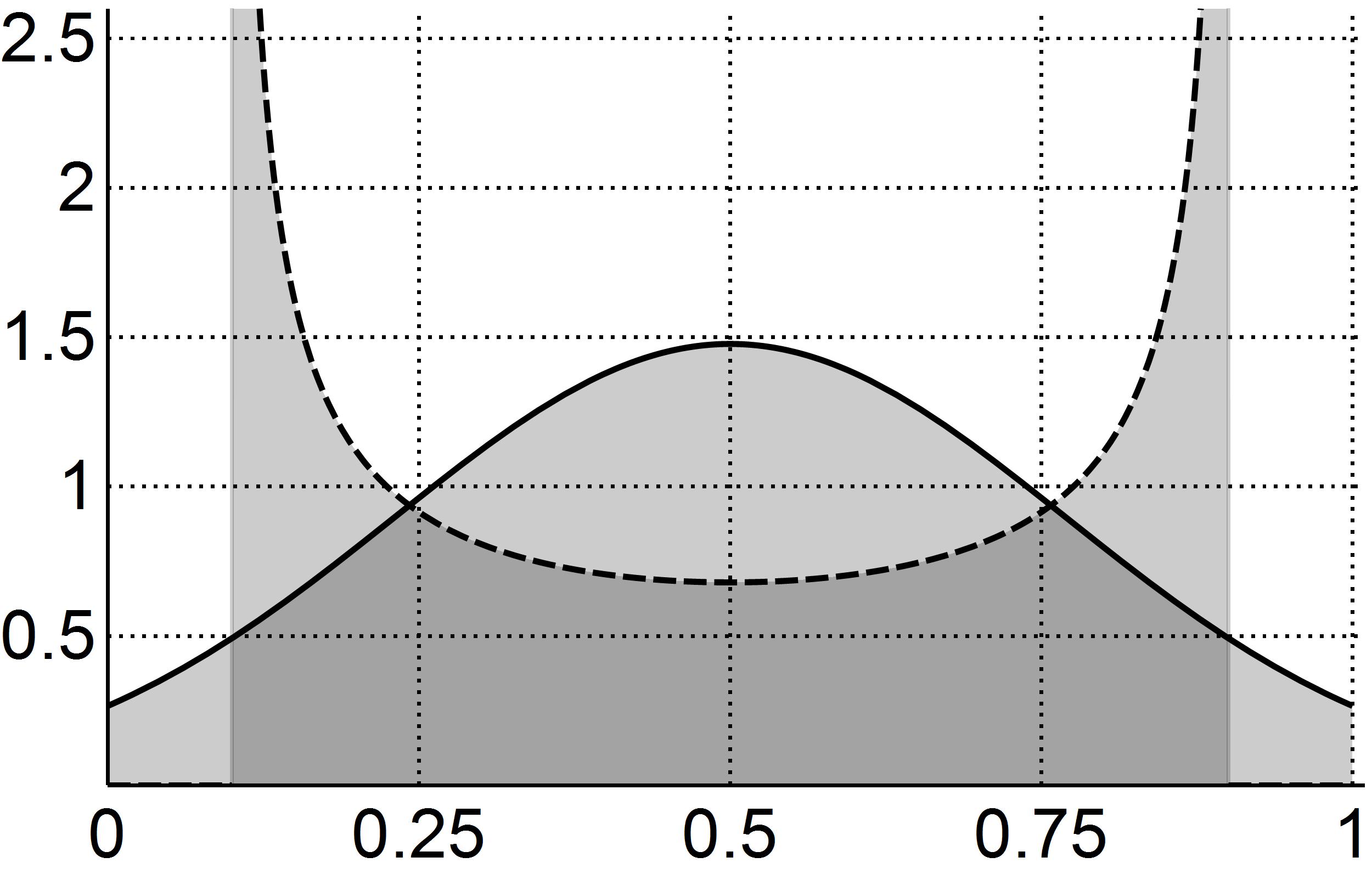}
			\end{subfigure}%
			~~~~~~~~
			\begin{subfigure}[b]{.40\textwidth}
				\centering
				\includegraphics[width=0.95\textwidth]{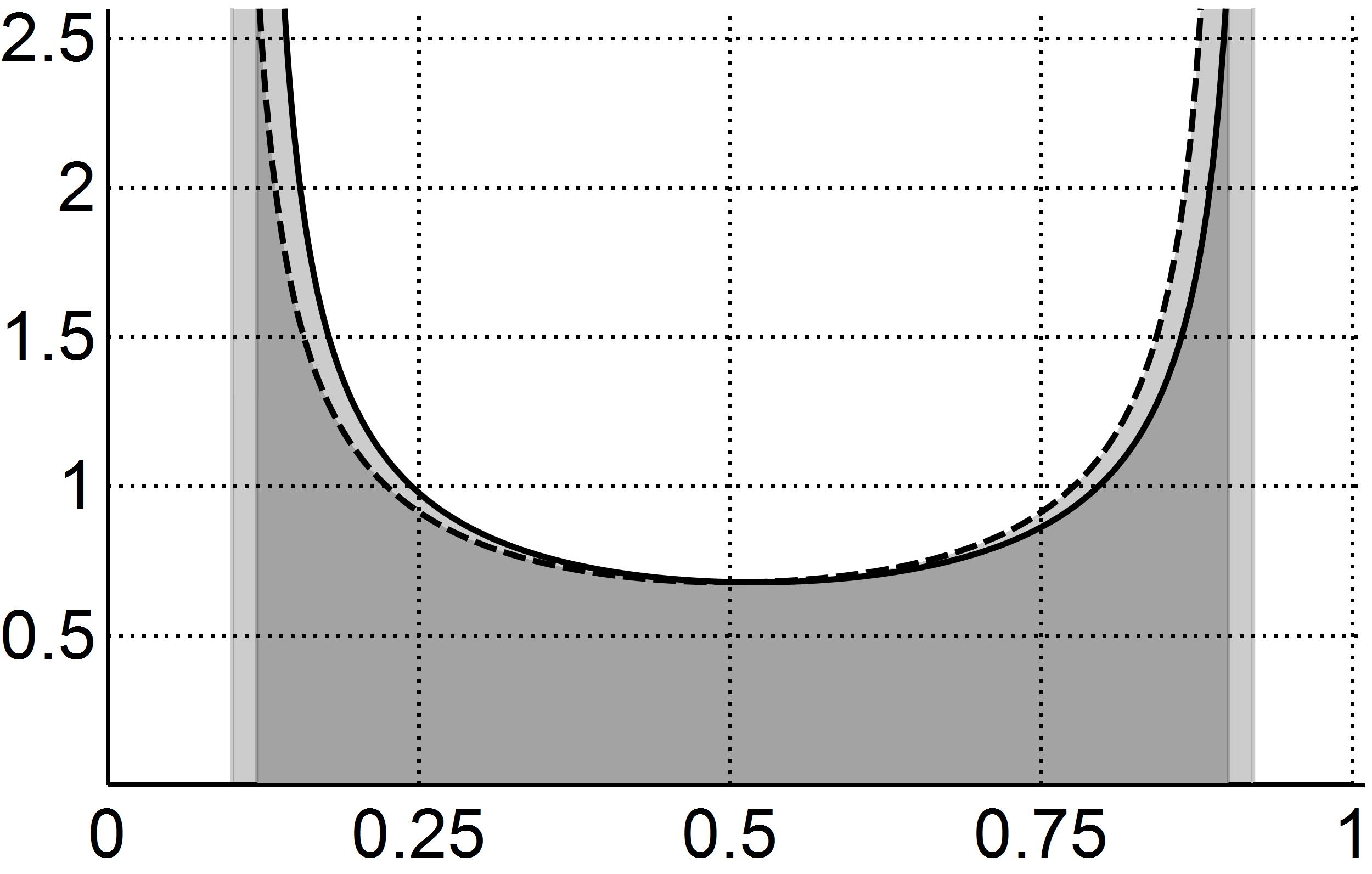}
			\end{subfigure}%
		}
		\caption{
			Illustrative example of the mean over-penalization problem.
			The MMD with standard polynomial kernel~\cite{gretton2006kernel} and different other raw moment based metrics~\cite{mroueh2017mcgan,li2017mmd} lead to counter-intuitive distance measurement as they consider the source Beta distribution (dashed) to be more similar to the Normal distribution on the left (solid) than to the slightly shifted Beta distribution on the right (solid).
			The proposed metric considers the distributions on the right to be more similar.
		}
		\label{fig:problem}
	\end{figure}

	For another example consider Fig.~\ref{fig:problem}.
	Different raw moment based metrics consider the source Beta distribution (dashed) to be more similar to the Normal distribution on the left (solid) than to the slightly shifted Beta distribution on the right (solid).
	This is especially the case for the integral probability metrics in Eq.\eqref{eq:ipm} with the polynomial spaces $\Pk^1$, $\Pk^2$ and $\Pk^4$, the MMD with the standard polynomial kernel $\kappa(x,y):=(1+\langle x,y\rangle_2)^2$ and the quartic kernel $\kappa(x,y):=(1+\langle x,y\rangle_2)^4$~\cite{gretton2006kernel,li2017mmd}, and the integral probability metrics in~\cite{mroueh2017mcgan}.
	See Section~\ref{app:penalization_problem} for the proof.
	
	Following first ideas as presented in~\cite{zellinger2017central}, we propose a metric that considers the distributions on the right to be more similar.
	
	\section{A Probability Metric for Distribution Alignment}
	\label{sec:cmd}
	Eq.\eqref{eq:meansensitive} motivates us to look for a modified version of the integral probability metric that is less sensitive to translation.
	Therefore, we  propose the following centralized and translation-invariant version of the integral probability metric between the distributions
	$\D$ and $\D'$:
	\begin{align}
		\begin{split}
			\label{eq:centralized_ipm}
			d_{\F}^c&(\D,\D') :=\\
			&\sup_{f\in\F}\big|\E_{\D}[f(\x-\E_{\D}[\x])]
			- \E_{\D'}[f(\x-\E_{\D'}[\x])]\big|\text{.}
		\end{split}
	\end{align}
	We apply this modification on our problem of domain adaptation by introducing a ``refined'' metric as the weighted sum of centralized integral probability metrics in Eq.\eqref{eq:centralized_ipm} with unit balls of polynomial reproducing kernel Hilbert spaces of different orders
	\begin{equation}
		\label{eq:cmd}
		\mathrm{cmd}_k(\D,\D') := a_1\, d_{\mathcal{P}^1}(\D,\D') + \sum_{j=2}^k  a_j\, d_{\Pk^j}^c(\D,\D')\text{,}
	\end{equation}
	where $a_j \geq 0$, $d_{\Pk}$ and $d_{\Pk}^c$ are defined as in Eq.\eqref{eq:ipm} and Eq.\eqref{eq:centralized_ipm}, respectively, \wrt the polynomial spaces $\Pk^{k}$.
	Note that in Eq.\eqref{eq:cmd} for $k=1$ we take $d_{\mathcal{P}^1}(\D,\D') = |\mu - \mu'|$ which still behaves smoothly \wrt changes of the mean values and is more informative than $d_{\mathcal{P}^1}^c(\D,\D') = 0$.
	
	The distance function in Eq.\eqref{eq:cmd} is a metric on the set of compactly supported distributions for $k=\infty$, and it is a pseudo-metric for $k<\infty$~\cite{zellinger2017central}.
	A zero value of this distance function implies equal moment characteristics. Therefore, it belongs  to the class of {\it primary} probability metrics also known as moment distances~\cite{rachev2013methods}.
	
	The questions of how to compute the metric efficiently, how to set the weighting values $a_j$ and how the minimization of the metric relates to the target error, are discussed in the next section.
	
	\section{Properties of the Probability Metric}
	\label{sec:properties}
	
	So far, our approach of defining an appropriate metric, i.e. Eq.\eqref{eq:cmd}, has been motivated by theoretical considerations starting from Eq.\eqref{eq:ipm} and the analysis in Section~\ref{sec:prob_metric}.
	However, for practical applications we need to compute our metric in a computationally efficient way.
	Theorem~\ref{thm:dual_cmd} provides a key step in this direction (see Section~\ref{app:dual} for its proof).
	\begin{thm}
		\label{thm:dual_cmd}
		By setting $c_1(\D)=\E_{\D}[\vec x]$ and 
		$c_j(\D)=\E_{\D}[\boldsymbol{\nu}^{(j)}(\vec x-\E_{\D}[\vec x])]$ for $j\geq2$ with the monomial vector as in Eq.\eqref{eq:nu}, 
		we obtain as  equivalent representation for the metric in Eq.\eqref{eq:cmd}:
		\begin{equation}
			\label{eq:dual_cmd}
			\mathrm{cmd}_k(\D,\D') = \sum_{j=1}^k a_j \norm{c_j(\D)-c_j(\D')}_2.
		\end{equation}
	\end{thm}
	Theorem~\ref{thm:dual_cmd} gives reason to call the metric in Eq.\eqref{eq:cmd} Central Moment Discrepancy.
	
	{In the special case of $k = 2$, the CMD in Eq.\eqref{eq:dual_cmd} is the weighted sum of the MMD with linear kernel and the Frobenius norm of the difference between the covariance matrices.
		This allows to interpret the CMD as an extension of correlation alignment approaches~\cite{sun2016deep,sun2016return} and linear kernel based MMD approaches~\mbox{\cite{tzeng2014deep}}.}
	
	The next practical aspect we must address is how to set the weighting factors $a_j$ in Eq.\eqref{eq:dual_cmd} such that the values of the terms of the sum do not increase too much.
	For distributions with compact support $[a,b]$, Proposition~\ref{prop:bound} provides us with suitable weighting factors, namely
	$$
	a_j:=1/{|b-a|^j}.
	$$
	\begin{prop}[Upper Central Moment Bound]
		\label{prop:bound}
		Let $\D$ and $\D'$ be two distributions supported on $[a,b]$ with finite mean values and $c_j$, $j=1,\ldots k$, as in Theorem~\ref{thm:dual_cmd}, then
		\begin{align}
			\begin{split}
				\frac{1}{|b-a|^j}&\norm{c_j(\D)-c_j(\D')}_2\\
				&\leq 2 \left(\frac{1}{j+1}\left(\frac{j}{j+1}\right)^j+\frac{1}{2^{1+j}}\right).
			\end{split}
		\end{align}
	\end{prop}
	Proposition~\ref{prop:bound} gives some insight into the contribution of lower and higher order central moment terms of the CMD in Eq.\eqref{eq:dual_cmd}.
	The upper bound strictly decreases with the order $j$ and shows that higher moment terms can contribute less than lower order moment terms to the overall value of \eqref{eq:dual_cmd}. See Section~\ref{app:upper} for the proof of Proposition~\ref{prop:bound}.
	
	It is natural to ask about the difference between the distributions $\D$ and $\D'$ given the value of $\mathrm{cmd}(\D,\D')$.
	This question is related to the problem of determining a distribution based on its moment sequence, also called {\it moment problem}.
	The moment problem can be uniquely solved for compactly supported distributions (Hausdorff moment problem).
	For distributions with different support, additional assumptions on the distributions are needed, \eg Carleman's condition (Hamburger moment problem, Stieltjes moment problem).
	Under such assumptions, the central moment discrepancy in Eq.\eqref{eq:cmd} can be used, together with an error term, to bound the absolute difference between characteristic functions and it therefore relates to weak convergence (see Section~\ref{app:characteristic} for the proofs).
	
	For simplicity let $\D_n$ for $n\in\mathbb{N}$ and $\D_\infty$ be distributions with support $[\nicefrac{-1}{2},\nicefrac{1}{2}]^m$, zero mean and finite moments of each order.
	Further, let $\zeta_n$ and $\zeta_\infty$ be the characteristic functions of $\D_n$ and $\D_\infty$, respectively.
	Then, $\sup_{\norm{\boldsymbol{t}}_1\leq 1}|\zeta_n(\boldsymbol{t}) - \zeta_\infty(\boldsymbol{t})|\rightarrow 0$
	entails weak convergence of the distributions $\D_n$ towards $\D_\infty$ and the following error bound holds.
	\begin{thm}[Characteristic Function Bound]
		\label{thm:ch_fct}
		For odd $k \in \mathbb{N}$ we have
		\begin{equation}
			\label{eq:ch_fct_bound}
			\begin{split}
				\sup_{\norm{\boldsymbol{t}}_1\leq 1}|\zeta_n(\boldsymbol{t}) &-
				\zeta_\infty(\boldsymbol{t})| \leq\\
				&\leq \sqrt{m}\, e\, \mathrm{cmd}_k(\tiny{\D_n,\D}) + \tau(\tiny{k,\D_n,\D}),
			\end{split}
		\end{equation}
		where
		\begin{equation}
			\tau(k,\D_n,\D) = \frac{1}{(k+1)!}\cdot \max_{\norm{\boldsymbol{\alpha}}_1=k+1}(|c_{\boldsymbol{\alpha}}(\D_n)|+|c_{\boldsymbol{\alpha}}(\D)|)
		\end{equation}
		and the ${\boldsymbol{\alpha}}$-moment of $\D$ is given by
		$c_{\boldsymbol{\alpha}}(\D)=\E_{\D}[x_1^{\alpha_1}\cdots x_m^{\alpha_m}]$ with $\boldsymbol{\alpha}=(\alpha_1,\ldots,\alpha_m)\in\mathbb{N}^m$.
	\end{thm}
	
	Theorem~\ref{thm:ch_fct} relates the minimization of the CMD to the minimization of the target error in 
	Theorem~\ref{thm:targeterror}.
	To see this, assume that $\tau(k,\D_n,\D)$ in Eq.\eqref{eq:ch_fct_bound} is zero.
	Then, convergence in CMD implies weak convergence of $\D_n$ to $\D$.
	Weak convergence is equivalent to convergence of $\E_{\D_n}[f]$ to $\E_{\D}[f]$ for all bounded continuous functions $f$.
	Therefore, if an algorithm forces $\mathrm{cmd}(\D_S,\D_T)$ to approach zero, it also forces the integral probability metric $d_\F(\D_S,\D_T)$ in Theorem~\ref{thm:targeterror} to approach zero for $\F$ being the class of all bounded continuous functions which is assumed to contain $|h-g_S|$ and $|h-g_T|$.
	Thus, Theorem~\ref{thm:targeterror} and Theorem~\ref{thm:ch_fct} together imply that the algorithm minimizes the target error.
	
	Note that, in the one-dimensional case, also lower bounds for Eq.\eqref{eq:ch_fct_bound} are known for primary probability distances~\cite[Theorem 10.3.6]{rachev2013methods}.
	
	So far, our analysis has been mainly theoretically motivated.
	In practice, not all cross-moments are always needed. 
	Our experiments show that reducing the monomial vector in Eq.\eqref{eq:nu} to
	\begin{equation}
		\label{eq:nu_marginal}
		\boldsymbol{\nu}^{(k)}(\vec x) := \left(x_1^k,\ldots,x_m^k\right)\text{.}
	\end{equation}
	leads already to better results than with comparable approaches while computational efficency is improved.

	\section{Domain Adaptation via Moment Alignment}
	\label{sec:MANN}
	
	We tackle the problem of minimizing the target error of a neural network by minimizing an approximation of the right side in the bound of 
	Theorem~\ref{thm:targeterror} by means of the minimization of the CMD in Eq.\eqref{eq:dual_cmd} between the domain-specific latent representations.
	For simplicity, we concentrate on the development of a minimization algorithm for a feed-forward neural network
	\begin{equation}
		\label{eq:nn}
		h = h_1\circ h_0:\mathbb{R}^m\times\Theta\rightarrow [0,1]^{|\mathcal{C}|}
	\end{equation}
	with parameter set $\Theta$ and a single hidden layer.
	The network maps input samples $X\subset\mathbb{R}^m$ to labels $Y\subset [0,1]^{|\mathcal{C}|}$, where $Y$ is an encoding of labels in $\mathcal{C}$.
	The first layer (hidden layer) $h_0:\mathbb{R}^m\times\Theta\rightarrow\mathbb{R}^n$ maps the inputs to the hidden activations $h_0(X)\in\mathbb{R}^n$.
	The second layer (classification layer) $h_1:\mathbb{R}^n\times\Theta\rightarrow [0,1]^{{|\mathcal{C}|}}$ maps the hidden activations to the labels.
	If it is clear from the application, we use the shorthand notation $h(\vec x)=h(\vec x;\Theta)$ and 
	$h(X)=\{h(\vec x)\}_{\vec x\in X}$ for the sample $X\subset\mathbb{R}^m$.
	
	As hidden layer, we use a standard fully connected layer with non-linear sigmoid activation function, i.e.
	\begin{equation}
		\label{eq:h0}
		h_0(\vec x) = h_0(\vec x;\vec{W},\vec{b}):=\mathrm{sigm}(\vec{W}\x + \vec b)
	\end{equation}
	with $\mathrm{sigm}(\x)=\Big(\frac{1}{1+e^{-x_1}},\ldots,\frac{1}{1+e^{-x_n}} \Big)$  and a matrix-vector parameter pair 
	$\Theta = (\vec W,\vec b)\in \mathbb{R}^{n\times m}\times\mathbb{R}^n$.
	
	The classification layer $h_1:\mathbb{R}^n\times\Theta\rightarrow [0,1]^{{|\mathcal{C}|}}$ is parametrized by
	$(\vec V,\vec c)\in \mathbb{R}^{{|\mathcal{C}|}\times n}\times\mathbb{R}^{|\mathcal{C}|}$ via
	\begin{equation}
		\label{eq:h1}
		h_1(\vec x;\vec{V},\vec{c}):=\mathrm{softmax}(\vec{V}\, h_0(\vec x) + \vec c)
	\end{equation}
	with $\mathrm{softmax}(\x)= (e^{x_1}, \ldots, e^{x_{|\mathcal{C}|}})\big/ \sum_{i=1}^{|\mathcal{C}|} e^{x_i} \in [0,1]^{|\mathcal{C}|}$ where the component-wise division is considered.
	Note that the softmax function enables the interpretation of the output $h(\x)=h_1(h_0(\x))$ as likelihood vector, i.e. the coordinate $h(\x)_i$ can be interpreted as the predicted likelihood that the vector $\x$ corresponds to the $i$-th label in $\mathcal{C}$.
	
	In the following we apply networks of the type defined in Eq.\eqref{eq:nn} to the problem of unsupervised domain adaptation as motivated in Section~\ref{sec:da}. Given a labeled source sample $(X_S,Y_S)\subset \mathbb{R}^m\times [0,1]^{|\mathcal{C}|}$ and an unlabeled target sample 
	$X_T\subset \mathbb{R}^m$, we want to train a classifier that performs well on unseen target data.
	As motivated in Section~\ref{sec:da_w_distr_al}, this problem can be tackled by training the neural network in Eq.\eqref{eq:nn} based on the objective
	\begin{align}
		\label{eq:obj}
		\begin{split}
			\min_{\vec W, \vec b,\vec V, \vec c} &\mathcal{L}(h_1(h_0(X_S;\vec W,\vec b);\vec V,\vec c),Y_S) \\
			&+ \lambda \cdot d(h_0(X_S;\vec W,\vec b),h_0(X_T;\vec W,\vec b))
		\end{split}
	\end{align}
	with an empirical loss $\mathcal{L}$ in the source domain and a distance function $d$ between the activations $h_0(X_S)$ and $h_0(X_T)$.
	See Fig.~\ref{fig:architecture} for an illustration.
	The parameter $\lambda$ is a trade-off parameter that articulates the priority of the domain adaptation compared to the source error minimization.
	Objective~\eqref{eq:obj} can be seen as a surrogate for the right side in Theorem~\ref{thm:targeterror}.
	
	A typical choice for the classification loss is the expectation of the negative log probability of the correct label
	\begin{equation}
		\label{eq:loss}
		\mathcal{L}(h(X_S),Y_S) := \frac{1}{|(X_S,Y_S)|}\sum_{(\x,\y)\in (X_S,Y_S)}l(h,\x,\y)
	\end{equation}
	with the estimator of the cross-entropy $l(h,\x,\y)=- \sum_{i=1}^{{|\mathcal{C}|}}y_i\log(h(\x)_i)$.
	
	We model the distance function $d$ in Eq.\eqref{eq:obj} by 
	an empirical estimate of the CMD in Eq.\eqref{eq:dual_cmd} based on
	\begin{align}
		\label{eq:cmd_estimate}
		\mathrm{cmd}(X_S,X_T) \sim \sum_{j=1}^{k} \norm{c_j(X_S)-c_j(X_T)}_2
	\end{align}
	for ${c_k(X)=\frac{1}{|X|}\sum_{\x\in X}\boldsymbol{\nu}^{(k)}(\x-c_1(X))}$ with  ${c_1(X)=\frac{1}{|X|}\sum_{\x\in X}\x}$ and $\boldsymbol{\nu}^{(k)}$ as defined in Eq.\eqref{eq:nu_marginal}.
	According to Proposition~\ref{prop:bound}, the weighting factors $a_j$ in Eq.\eqref{eq:dual_cmd} are set to one as the sigmoid function maps to the interval $[0,1]$.
	Note that the estimate in Eq.\eqref{eq:cmd_estimate} is consistent but biased~\cite{zellinger2017central}.
	To obtain an unbiased estimate of a moment distance with similar properties as the CMD in Eq.\eqref{eq:dual_cmd}, we can apply the sample central moments as unbiased estimates of the central moments and use the squared Euclidean norm instead of the Euclidean norm in Eq.\eqref{eq:dual_cmd} as similarly proposed for the MMD~\cite{gretton2006kernel}.
	
	We tackle the optimization of Eq.\eqref{eq:obj} by stochastic gradient descent.
	Let the objective function be
	\begin{align}
		\label{eq:obj_approx}
		J(\Theta) &:=\mathcal{L}(h(X_S;\Theta),Y_S) + \lambda\cdot \mathrm{cmd}(X_S,X_T)\text{.}
	\end{align}
	with the negative log probability $\mathcal{L}(h(X;\Theta),Y)$ as in Eq.\eqref{eq:loss} and the CMD estimate as in Eq.\eqref{eq:cmd_estimate}.
	Then, the gradient update step is given by
	\begin{align}
		\label{eq:update}
		\Theta^{(k+1)} &:= \Theta^{(k)} - \alpha\cdot \eta^{(k)} \cdot \nabla_\Theta J(\Theta^{(k)}) ,
	\end{align}
	with learning rate $\alpha$ and gradient weighting $\eta^{(k)}$.
	$\nabla_\Theta$ denotes the gradient \wrt~$\Theta$.
	The gradients of Eq.\eqref{eq:obj_approx} are derived in Section~\ref{app:gradients}.
	
	In the case of sparse data as in the sentiment analysis experiments in Section~\ref{subsec:amazon_reviews}, we rely on Adagrad~\cite{duchi2011adaptive} gradient weighting
	\begin{align}
		\label{eq:adagrad}
		\begin{split}
			\eta^{(k)} &:= \frac{1}{\sqrt{G^{(k)}}}\\
			G^{(k+1)} &:= G^{(k)} + (\nabla_\Theta J(\Theta^{(k)}))^2
		\end{split}
	\end{align}
	where the division and the square root are taken element-wise.
	Eq.\eqref{eq:adagrad} can be interpreted as gradient update according to different update weights for each dimension, i.e. the weighting parameter $\alpha$ is divided by the norm of the historical gradient separately for each hidden node.
	The idea is to give frequently occurring features very low learning rates and infrequent features high learning rates.
	
	In the case of non-sparse data, as in the experiments on artificial data in Section~\ref{subsec:artificial_dataset} and in the experiments on image data in Section~\ref{subsec:office}, we use the Adadelta weighting scheme
	\begin{align}
		\label{eq:adadelta}
		\begin{split}
			G^{(k)} &:= \rho G^{(k-1)} + (1-\rho) (\nabla_\Theta J(\Theta^{(k)}))^2\\
			\eta^{(k)} &:= \frac{\sqrt{E^{(k-1)}+\epsilon}}{\sqrt{G^{(k)}}}\\
			E^{(k)} &:= \rho E^{(k-1)} - (1-\rho) (\eta^{(k-1)}\cdot \nabla_\Theta J(\Theta^{(k)}) )^2,
		\end{split}
	\end{align}
	where $\rho$ is a decay constant and $\epsilon$ is a small number for numerical stability.
	The Adadelta gradient weighting scheme in Eq.\eqref{eq:adadelta} is an extension of the Adagrad gradient weighting scheme in Eq.\eqref{eq:adagrad} that seeks to reduce its aggressive, monotonically decreasing learning rate by considering also historical gradient updates $E^{(k)}$.
	Adadelta requires no manual tuning of a learning rate, i.e. $\alpha=1$, and appears robust to noisy gradient information, different model architecture choices, various data modalities and selection of hyper-parameters.
	Adadelta is therefore a suitable choice for our aim of creating a robust learning algorithm.
	
	{As noted in Section~\ref{s:intro}, the tuning of the domain adaptation weight $\lambda$ is sophisticated without labels in the target domain.
		In order to increase the visibility of the effects of the proposed method we refrain from hyper parameter tuning but carry out our experiments with a fixed parameter set.}
	In order to articulate our preference to treat both terms as equally important, it is therefore reasonable to set $\lambda=1$.
	{However, this leaves space for additional improvement of model accuracies via the development of unsupervised parameter tuning techniques.}
	
	\SetKwInOut{Stepzero}{Step 0}
	\SetKwInOut{Stepone}{Step 1}
	\SetKwInOut{Steptwo}{Step 2}
	\SetKwInOut{Stepthree}{Step 3}
	\SetKwInOut{Init}{Init}
	\begin{algorithm}
		\small
		\label{alg:mann}
		\SetAlgoLined
		\KwIn{Samples $(X_S,Y_S)\subset \mathbb{R}^m\times [0,1]^{|\mathcal{C}|}$ and $X_T\subset \mathbb{R}^m$
		}
		\KwOut{Neural network parameters $\{\vec{W}, \vec b, \vec V, \vec c\}$}~\\
		\Init{Initialize parameters $\vec{W}$, $\vec b$, $\vec V$ and $\vec c$ randomly.}
		\While{\upshape stopping criteria is not met}{
			\Stepone{Compute the source activations $h_0(X_S;\vec W,\vec b)$, the target activations $h_0(X_T;\vec W,\vec b)$ and the source outputs $h(X_S;\vec W,\vec b, \vec V, \vec c)$ according
				to Eq.\eqref{eq:h0} and Eq.\eqref{eq:h1}.}
			\Steptwo{Compute the gradients of Eq.\eqref{eq:obj_approx} \wrt $\vec{W}$, $\vec b$, $\vec V$ and $\vec c$ as in Section~\ref{app:gradients}.}
			\Stepthree{Update the parameters $\vec{W}$, $\vec b$, $\vec V$ and $\vec c$ according to Eq.\eqref{eq:update}.}
		}
		\caption{Moment Alignment Neural Network - Stochastic Gradient Update}
	\end{algorithm}
	
	Let $n$ be the number of hidden nodes of the network, then
	the gradient update in Step 2 can be implemented with linear time complexity $\mathcal{O}\left(n\cdot(|X_S|+|X_T|)\right)$ by the formulas derived in Section~\ref{app:gradients}.
	Note that this is an improvement over MMD-based approaches (which compute the full kernel matrix) and correlation matrix alignment approaches in terms of computational complexity which is $\mathcal{O}\left(n\cdot(|X_S|^2+|X_S|\cdot |X_T|+|X_T|^2)\right)$ for MMD and $\mathcal{O}\left( n\cdot|X_S|\cdot|X_T| \right) $ for correlation alignment approaches.
	
	\section{Experiments}
	\label{sec:experiments}
	
	Our experimental evaluations are based on seven datasets, one artificial dataset, two benchmark datasets for domain adaptation, {\it Amazon reviews} and {\it Office} and four digit recognition datasets, {\it MNIST}, {\it SVHN}, {\it MNIST-M} and {\it SynthDigits}. They are described in Subsection~\ref{subsec:datasets}.
	
	Our experiments aim at providing evidence regarding the following aspects: On the usefulness of our algorithm for adapting neural networks to artificially shifted and rotated data (Subsection~\ref{subsec:artificial_dataset}), on the classification accuracy of the proposed algorithm on the sentiment analysis of product reviews based on the learning of neural networks with a single hidden-layer (Subsection~\ref{subsec:amazon_reviews}), On the classification accuracy on object recognition tasks based on the learning of pre-trained convolutional neural networks (Subsection~\ref{subsec:office}), on the classification accuracy of deep convolutional neural networks trained on raw image data (Subsection~\ref{subsec:image_rec}), and, on the accuracy sensitivity regarding changes in the number-of-moments parameter and changes in the number of hidden nodes (Subsection~\ref{subsection:sensitivity}).
	
	\subsection{Datasets}
	\label{subsec:datasets}
	
	The following datasets are summarized in Table~\ref{tab:datasets}.
	
	{\bf Artificial dataset:}
	In order to analyze the applicability of our algorithm for adapting neural networks to rotated and shifted data, we created an artificial dataset (Fig.~\ref{fig:artificial_problem}).
	The source data consists of three classes that are arranged in the two-dimensional space.
	Different transformations such as shifts and rotations are applied on all classes to create unlabeled target data.
	
	{\bf Sentiment analysis:}
	To analyze the accuracy of the proposed approach on sentiment analysis of product reviews, we rely on the {\it Amazon reviews} benchmark dataset with the same preprocessing as used by others~\cite{chen2012marginalized,ganin2016domain,louizos2015variational}.
	The dataset contains product reviews of four categories: {\it books} (B), {\it DVDs} (D), {\it electronics} (E) and {\it kitchen appliances} (K). Reviews are encoded in 5000 dimensional feature vectors of bag-of-words unigrams and bigrams with binary labels: label $0$ if the product is ranked by $1$ to $3$ stars and label $1$ if the product is ranked by $4$ or $5$ stars. From the four categories we obtain twelve domain adaptation tasks where each category serves once as source domain and once as target domain.
	
	{\bf Object recognition:}
	In order to analyze the accuracy of our algorithm on an object recognition task, we perform experiments based on the {\it Office} dataset~\cite{saenko2010adapting}, which contains images from three distinct domains: {\it amazon} (A), {\it webcam} (W) and {\it DSLR} (D).
	This dataset is a well known benchmark for domain adaptation algorithms in computer vision.
	According to the standard protocol~\cite{ganin2016domain,long2015learning}, we downsample and crop the images such that all are of the same size $({227\times 227})$.
	We assess the performance of our method across all six possible transfer tasks.
	
	{\bf Digit recognition:}
	To analyze the accuracy of our algorithm on digit recognition tasks, we rely on domain adaptation between the three digit recognition datasets {\it MNIST}~\cite{lecun1998mnist}, {\it SVHN}~\cite{netzer2011reading}, {\it MNIST-M}~\cite{ganin2016domain} and {\it SynthDigits}~\cite{ganin2016domain}.
	MNIST contains $70000$ black and white digit images, SVHN contains $99289$ images of real world house numbers extracted from Google Street View and MNIST-M contains $59001$ digit images created by using the MNIST images as a binary mask and inverting the images with the colors of a background image. The background images are random crops uniformly sampled from the Berkeley Segmentation Data Set.
	SynthDigits contains $500000$ digit images generated by varying the text, positioning, orientation, background, stroke colors and blur of Windows$^\text{TM}$ fonts.
	According to the standard protocol~\cite{tzeng2017adversarial}, we resize the images $({32\times 32})$.
	We compare our method based on the standard benchmark experiments SVHN$\rightarrow$MNIST and MNIST$\rightarrow$MNIST-M (source$\rightarrow$target).
	The datasets are summarized in Table~\ref{tab:datasets}.
	
	\begin{table}[!h]
		\renewcommand{\arraystretch}{1.3}
		\scriptsize
		\centering
		\begin{tabular}{|c|c||c|c|c|c|}
			\hline
			\bfseries Task & Domain/Dataset & \bfseries Samples & \bfseries Classes & \bfseries Features\\
			\hline\hline
			\multirow{2}{*}{\begin{tabular}{@{}c@{}}Artificial\\example\end{tabular}} & Source & $639$ & $3$ & $2$\\
			\cline{2-5}
			& Target & $639$ & $3$ & $2$\\
			\hline
			\hline
			\multirow{4}{*}{\begin{tabular}{@{}c@{}}Sentiment\\analysis\end{tabular}} & Books (B) & $6465$ & $2$ & $5000$\\
			\cline{2-5}
			& DVDs (D) & $5586$ & $2$ & $5000$\\
			\cline{2-5}
			& Electronics (E) & $7231$ & $2$ & $5000$\\
			\cline{2-5}
			& Kitchen appliances (K) & $7945$ & $2$ & $5000$\\
			\hline
			\hline
			\multirow{3}{*}{\begin{tabular}{@{}c@{}}Object\\recognition\end{tabular}} & Amazon (A) & $2817$ & $31$ & $227\times 227$\\
			\cline{2-5}
			& Webcam (W) & $795$ & $31$ & $227\times 227$\\
			\cline{2-5}
			& DSLR (D) & $498$ & $31$ & $227\times 227$\\
			\hline
			\hline
			\multirow{4}{*}{\begin{tabular}{@{}c@{}}Digit\\recognition\end{tabular}} & SVHN & $99289$ & $10$ & $32\times 32$\\
			\cline{2-5}
			& MNIST & $70000$ & $10$ & $32\times 32$\\
			\cline{2-5}
			& MNIST-M & $59001$ &  $10$ & $32\times 32$\\
			\cline{2-5}
			& SynthDigits & $500000$ &  $10$ & $32\times 32$\\
			\hline
		\end{tabular}
		\caption{Datasets}
		\label{tab:datasets}
	\end{table}
	
	\subsection{Artificial Example}
	\label{subsec:artificial_dataset}
	
\begin{figure}[!h]
		\makebox[\linewidth][c]{%
			\begin{subfigure}[b]{.45\textwidth}
				\centering
				\includegraphics[width=.95\textwidth]{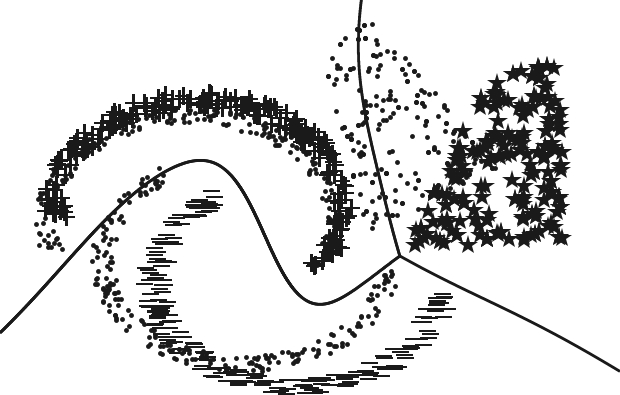}
			\end{subfigure}%
			\begin{subfigure}[b]{.45\textwidth}
				\centering
				\includegraphics[width=0.95\textwidth]{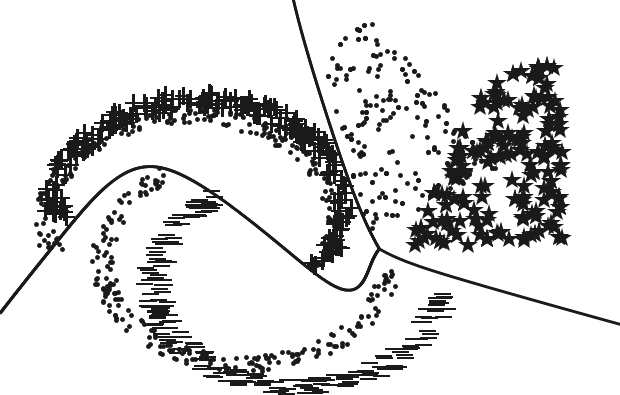}
			\end{subfigure}%
		}
		\caption{Artificial classification scenario with three classes \mbox{("$+$", "$-$" and stars)} in the source domain and unlabeled data in the target domain (points) solved by Algorithm~1. Left: without domain adaptation, i.e. without the $\mathrm{cmd}$-term in Step~2; Right: with the proposed approach.}
		\label{fig:artificial_problem}
	\end{figure}

	The artificial dataset is described in Section~\ref{subsec:datasets} and visualized in Fig.~\ref{fig:artificial_problem}.
	We study the adaptation capability of our algorithm by comparing it to a standard neural network as described in Section~\ref{sec:MANN} with $15$ hidden neurons.
	That is, we apply Algorithm~1 twice, once without the CMD in Eq.\eqref{eq:obj_approx} and once with the CMD term.
	We refer to the two versions as shallow neural network (shallow NN) and moment alignment neural network (MANN) respectively.
	To start from a similar initial situation, we use the weights of the shallow NN after $\nicefrac{2}{3}$ of the training time as initial weights for the MANN and train the MANN for $\nicefrac{1}{3}$ of the training time of the shallow NN.
	
	The classification accuracy of the shallow NN in the target domain is $86.7\%$ and the accuracy of the MANN is $99.7\%$.
	The decision boundaries of the algorithms are shown in Fig.~\ref{fig:artificial_problem}, shallow NN on the left and MANN on the right.
	The shallow NN misclassifies some data points of the "$+$"-class and of the star-class in the target domain (points).
	The MANN clearly adapts the decision boundaries to the target domain and only a small number of points ($0.3\%$) is misclassified.
	We recall that this is the founding idea of our algorithm.
	
\begin{figure}
		\makebox[\linewidth][c]{%
			\begin{subfigure}[b]{.9\textwidth}
				\centering
				\includegraphics[width=.95\textwidth]{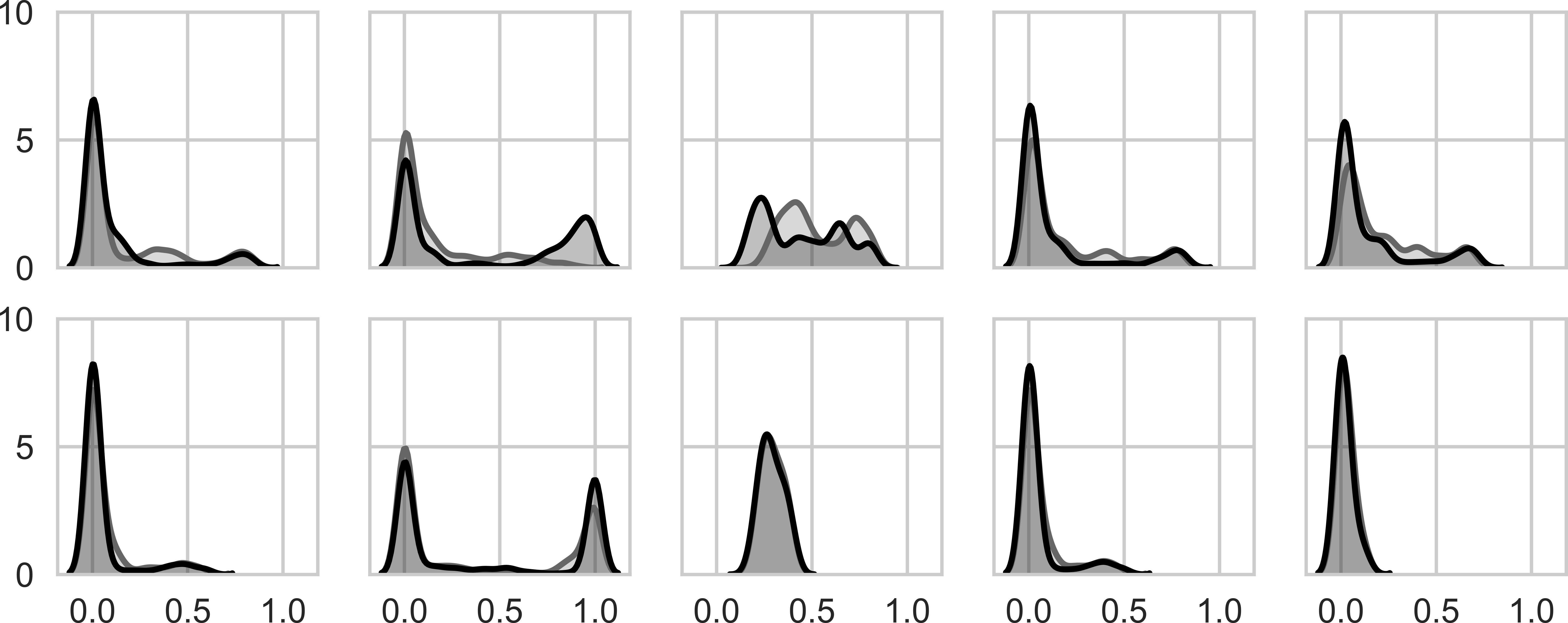}
			\end{subfigure}
		}
		\caption{Five most different source (dark gray) and target (light gray) activation distributions of the hidden nodes of the neural networks trained by Algorithm~1 on the artificial dataset (Fig.~\ref{fig:artificial_problem}) without domain adaptation (top) and with the proposed approach (bottom).}
		\label{fig:artificial_activations}
	\end{figure}

	Let us now test the hypothesis that the CMD helps to align the activation distributions of the hidden nodes.
	We measure the significance of a distribution difference by means of the p-value of a two-sided Kolmogorov-Smirnov test for goodness of fit.
	For the shallow NN, $13$ out of $15$ hidden nodes show significantly different distributions, whereas for the MANN only five distribution pairs are considered as being significantly different (p-value lower than $10^{-2}$).
	Kernel density estimates of these five distribution pairs are visualized in Fig.~\ref{fig:artificial_activations} (bottom).
	Fig.~\ref{fig:artificial_activations} (top) shows kernel density estimates of the distribution pairs corresponding to the five smallest p-values of the shallow NN.
	As the only difference between the two algorithms is the CMD, we conclude that the CMD successfully helps to align the activation distributions in this example.
	
	\subsection{Sentiment Analysis of Product Reviews}
	\label{subsec:amazon_reviews}
	
	In the following experiment, we compare our algorithm to related approaches based on the single-layer neural network architecture proposed in Section~\ref{sec:MANN}.
	
	We use the Amazon reviews dataset with the same data splits as previous works for every task~\cite{chen2012marginalized,louizos2015variational,ganin2016domain}. Thus, we have $2000$ labeled source examples and $2000$ unlabeled target examples for training, and between $3000$ and $6000$ examples for testing.
	
	Since no target labels are available in the unsupervised domain adaptation setting, we cannot select parameters via standard cross-validation procedures.
	Therefore, we apply a variant of the {\it reverse validation} approach~\cite{zhong2010cross} as refined for neural networks in~\cite{ganin2016domain}.
	
	We report results for representatives of all three research directions described in Section~\ref{sec:related} and one kernel learning method:
	\begin{itemize}
		\item {\it Shallow Neural Network (NN)}: Trained by Algorithm~1 without domain adaptation ($\lambda=0$ in Eq.\eqref{eq:obj_approx}) on the neural network architecture of Section~\ref{sec:MANN} with $50$ hidden nodeso~\cite{ganin2016domain}.
		
		\item {\it Transfer Component Analysis (TCA)}~\cite{pan2011domain}: This kernel learning algorithm tries to learn some transfer components across domains in an RKHS using the MMD. For competitive classification accuracies, we report results from~\cite{li2017prediction} where the optimal model architecture is searched in a supervised manner by also considering target labels.
		The trade-off parameter of the TCA is set to $\mu=0.1$ and the optimal dimension of the subspace is searched for $k\in\{10,20,\ldots,100,500\}$.
		
		\item {\it Domain-Adversarial Neural Networks (DANN)}~\cite{ganin2016domain}: This algorithm is summarized in Section~\ref{sec:related}. We report the results of the original paper, where the adaptation weighting parameter $\lambda$ is chosen among $9$ values between $10^{-2}$ and 1 on a logarithmic scale.
		The hidden layer size is either $50$ or $100$ and the learning rate is set to $10^{-3}$.
		
		\item {\it Deep Correlation Alignment (Coral)}~\cite{sun2016deep}: We apply Algorithm~1 with the CORAL distance function instead of the CMD in Eq.\eqref{eq:obj_approx}.
		We use the default parameter $\lambda=1$ as suggested in the original paper~\cite{sun2016deep}.
		
		\item {\it Maximum Mean Discrepancy (MMD)}~\cite{gretton2006kernel}: We apply Algorithm~1 using the MMD with Gaussian kernel instead of the CMD in Eq.\eqref{eq:obj_approx}.
		Parameter $\lambda$ is chosen among $10$ values between $0.1$ and $500$ on a logarithmic scale.
		The Gaussian kernel parameter is chosen among $10$ values between $0.01$ and $10$ on a logarithmic scale.
		
		\item {\it Central Moment Discrepancy (CMD)}:
		In order to increase the visibility of the effects of the proposed method we refrain from hyper parameter tuning but carry out our experiments with the same fixed parameter values of $\lambda$ and $k$ for all experiments.
		The number-of-moments parameter $k$ of the CMD in Eq.\eqref{eq:cmd_estimate} is heuristically set to five, as the first five moments capture rich geometric information about the shape of a distribution and $k=5$ is small enough to be computationally efficient.
		Note that the experiments in Section~\ref{subsection:sensitivity} show that similar results are obtained for all $k\in\{4,\ldots,7\}$.
		We use the default parameter $\lambda=1$ to articulate our preference that domain adaptation is equally important as the classification accuracy in the source domain.
	\end{itemize}
	Since we deal with sparse data, we rely on Adagrad~\cite{duchi2011adaptive} optimization technique in Eq.\eqref{eq:adagrad}.
	For all evaluations, the default parametrization is used as implemented in {\it Keras}~\cite{chollet2015keras}.
	We repeat our experiments ten times with different random initializations.
	
	The mean values and average ranks over all tasks are shown in Table~\ref{tab:am_rev_results}.
	Our method outperforms others in average accuracy as well as in average rank in all except one task.
	
	\begin{table}[!ht]
		\scriptsize
		\renewcommand{\arraystretch}{1.3}
		\centering
		\begin{tabular}{|c||c|c|c|c|c||c|}
			\hline
			Method 	&			NN 	&			DANN~\cite{ganin2016domain} 	&			CORAL~\cite{sun2016deep} 	&			TCA~\cite{pan2011domain} 	&			MMD~\cite{gretton2006kernel} 	&				CMD (ours) 		\\
			\hline
			\hline
			B$\shortrightarrow$D 	&	 $78.7$ 	&	 $78.4$ 	&	 $79.2$ 	&	 $78.9$ 	&	 $\mathit{79.6}$ 	&		 $\mathbf{80.5}$ 		\\
			\hline
			B$\shortrightarrow$E 	&	 $71.4$ 	&	 $73.3$ 	&	 $73.1$ 	&	 $74.2$ 	&	 $75.8$ 	&		 $\mathbf{78.7}$ 		\\
			\hline
			B$\shortrightarrow$K 	&	 $74.5$ 	&	 $77.9$ 	&	 $75.0$ 	&	 $73.9$ 	&	 $\mathit{78.7}$ 	&		 $\mathbf{81.3}$ 		\\
			\hline
			D$\shortrightarrow$B	&	 $74.6$ 	&	 $72.3$ 	&	 $77.6$ 	&	 $77.5$ 	&	 $\mathit{78.0}$ 	&		 $\mathbf{79.5}$ 		\\
			\hline
			D$\shortrightarrow$E	&	 $72.4$ 	&	 $75.4$ 	&	 $74.9$ 	&	 $\mathit{77.5}$ 	&	 $76.6$ 	&		 $\mathbf{79.7}$ 		\\
			\hline
			D$\shortrightarrow$K 	&	 $76.5$ 	&	 $78.3$ 	&	 $79.2$ 	&	 $\mathit{79.6}$ 	&	 $\mathit{79.6}$ 	&		 $\mathbf{83.0}$ 		\\
			\hline
			E$\shortrightarrow$B	&	 $71.1$ 	&	 $71.3$ 	&	 $71.6$ 	&	 $72.7$ 	&	 $\mathit{73.3}$ 	&		 $\mathbf{74.4}$ 		\\
			\hline
			E$\shortrightarrow$D	&	 $71.9$ 	&	 $73.8$ 	&	 $72.4$ 	&	 $\mathit{75.7}$ 	&	 $74.8$ 	&		 $\mathbf{76.3}$ 		\\
			\hline
			E$\shortrightarrow$K 	&	 $84.4$ 	&	 $85.4$ 	&	 $84.5$ 	&	 $\mathbf{86.6}$ 	&	 $85.7$ 	&		 $\mathit{86.0}$ 		\\
			\hline
			K$\shortrightarrow$B	&	 $69.9$ 	&	 $70.9$ 	&	 $73.0$ 	&	 $71.7$ 	&	 $\mathit{74.0}$ 	&		 $\mathbf{75.6}$ 		\\
			\hline
			K$\shortrightarrow$D	&	 $73.4$ 	&	 $74.0$ 	&	 $75.3$ 	&	 $74.1$ 	&	 $\mathit{76.3}$ 	&		 $\mathbf{77.5}$ 		\\
			\hline
			K$\shortrightarrow$E 	&	 $83.3$ 	&	 $84.3$ 	&	 $84.0$ 	&	 $83.5$ 	&	 $\mathit{84.4}$ 	&		 $\mathbf{85.4}$ 		\\
			\hline
			\hline
			Average 	&	 $75.2$ 	&	 $76.3$ 	&	 $76.7$ 	&	 $77.2$ 	&	 $\mathit{78.1}$ 	&		 $\mathbf{79.8}$ 		\\
			\hline
			Average rank	&	 $5.8$	&	 $4.5$	&	 $4.0$	&	 $3.3$	&	 $\mathit{2.3}$	&		 $\mathbf{1.1}$\\
			\hline
		\end{tabular}
		\caption{Classification accuracy on Amazon reviews dataset for twelve domain adaptation scenarios (source$\shortrightarrow$target)}
		\label{tab:am_rev_results}
	\end{table}
	
	\subsection{Object Recognition}
	\label{subsec:office}
	
	In the following experiments, we investigate our approach based on the learning of a pre-trained deep convolutional neural network. We aim at a robust approach, i.e. we try to find a balance between a low number of parameters and a high accuracy.
	
	Since the Office dataset is rather small (with only $2817$ images in its largest domain), we employ the pretrained convolutional neural network {\it AlexNet}~\cite{krizhevsky2012imagenet}.
	We follow the standard training protocol for this dataset and use the fully labeled source sample and the unlabeled target sample for training~\cite{long2015learning,ganin2016domain,sun2016deep,long2016unsupervised,long2016joint} and the target labels for testing.
	Using this "fully-transductive" protocol, we compare the proposed approach to the most related distribution alignment methods as described in Section~\ref{subsec:amazon_reviews}.
	For a fair comparison we report original results of works that only align the distributions of a single neural network layer of the AlexNet after the layer called $fc7$.
	
	We compare our algorithm to the following approaches:
	\begin{itemize}
		\item {\it Convolutional Neural Network (CNN)}~\cite{krizhevsky2012imagenet}: We apply Algorithm~1 without domain adaptation ($\lambda=0$ in Eq.\eqref{eq:obj_approx}) to the network architecture of Section~\ref{sec:MANN} on top of the output of the {\it fc7}-layer of AlexNet.
		We use a hidden layer size of $256$~\cite{tzeng2014deep,ganin2016domain}.
		Following the suggestions in~\cite{sun2016deep,ganin2016domain,long2015learning}, we randomly crop and mirror the images, ensure a balanced source batch and optimize using stochastic gradient descent with a momentum term of $0.9$ and a learning rate decay.
		In order to increase the visibility of the effects of the proposed method we refrain from hyper parameter tuning but carry out our experiments with the Keras~\cite{chollet2015keras} default learning rate and a decay of $10^{-4}$.
		
		\item {\it Transfer Component Analysis (TCA)}~\cite{pan2011domain}: We report results~\cite{long2016joint} that are based on the output of the {\it fc7}-layer of AlexNet with parameters tuned via reverse validation~\cite{zhong2010cross}.
		
		\item {\it Domain-Adversarial Neural Networks (DANN)}~\cite{ganin2016domain}: The original paper~\cite{ganin2016domain} reports results for the adaptation tasks A$\shortrightarrow$W, D$\shortrightarrow$W and W$\shortrightarrow$D.
		For the rest of the scenarios, we report the results of~\cite{long2016joint}.
		The distribution alignment is based on a $256$-sized layer on top of the $fc7$-layer.
		The images are randomly cropped and mirrored and stochastic gradient descent is applied with a momentum term of $0.9$.
		The learning rate is decreased polynomially and divided by ten for the lower layers.
		It is proposed to decrease the weighting parameter $\lambda$ in Eq.\eqref{eq:obj} with exponential order according to a specifically designed $\lambda$-schedule~\cite{ganin2016domain}.
		
		\item {\it Deep Correlation Alignment (CORAL)}~\cite{sun2016deep}: We report the results and parameters of the original paper in which they perform domain adaptation on a $31$-sized layer on top of the $fc7$-layer.
		Stochastic gradient descent is applied with a learning rate of $10^{-3}$, weight decay of $5\cdot 10^{-4}$ and momentum of $0.9$.
		The domain adaptation weighting parameter $\lambda$ is chosen in a way that "at the end of training the classification loss and the CORAL loss are roughly the same"~\cite{sun2016deep}.
		
		\item {\it Maximum Mean Discrepancy (MMD)}~\cite{gretton2006kernel}: We report the results of Long et al.~\cite{long2015learning} in which the MMD is applied on top of the $31$-dimensional layer after the $fc7$-layer.
		The domain adaptation weighting parameter $\lambda$ is chosen based on assessing the error of a two-sample classifier according to~\cite{fukumizu2009kernel}.
		A multi-kernel version of the MMD is used with varying bandwidth of the Gaussian kernel between $2^{-8}\gamma$ and $2^{8}\gamma$ with a multiplicative step-size of $\sqrt{2}$.
		Parameter $\gamma$ is chosen as the mean pairwise distance on the training data following the {\it median heuristic}~\cite{gretton2012optimal}.
		The network is trained via stochastic gradient descent with a momentum of $0.9$, polynomial learning rate decay and cross-validated initial learning rate between $10^{-5}$ and $10^{-2}$ with a multiplicative step size of $\sqrt{10}$.
		The learning rate is set to zero for the first three layers and for the lower layers it is divided by $10$.
		The images are randomly cropped and mirrored in this approach to stabilize the learning process.
		
		\item {\it Central Moment Discrepancy (CMD)}: The approach of this paper with the same optimization strategy as in CNN, with the number-of-moments parameter $k=5$ and the domain adaptation weight $\lambda=1$ as described in Section~\ref{subsec:amazon_reviews}.
		
		\item {\it Few Parameter Central Moment Discrepancy (FP-CMD)}:
		This approach aims at a low number of parameters.
		The Adadelta gradient weighting scheme (Eq.\eqref{eq:adadelta}) is used in the method above instead of the momentum.
		In addition, no data augmentation is applied.
	\end{itemize}
	
	The parameter settings of the neural network based approaches are summarized in Table~\ref{tab:param_settings}.
	
	\begin{table}[!ht]
		\renewcommand{\arraystretch}{1.3}
		\scriptsize
		\centering
		\begin{tabular}{|c||c|c|c||c|c|}
			\hline
			Method 		&			CORAL~\cite{sun2016deep} 	&			DANN~\cite{ganin2016domain} 	&			MMD~\cite{long2015learning} 		&			CMD (ours) 	&			FP-CMD (ours) 	\\
			\hline
			\hline
			\begin{tabular}{@{}c@{}}Adaptation \\ nodes\end{tabular} 		&	 $\mathbf{31}$ 	&	 $256$ 	&	 $\mathbf{31}$ 		&	 $256$ 	&	 $256$ 	\\
			\hline
			\begin{tabular}{@{}c@{}}Adaptation \\ weight $\lambda$\end{tabular} 		&	 \begin{tabular}{@{}c@{}}manually \\ tuned\end{tabular} 	&	 exp. decay 	&	 class. strategy 		&	 $\mathbf{1.0}$ 	&	 $\mathbf{1.0}$ 	\\
			\hline
			\begin{tabular}{@{}c@{}}Additional \\ hyper-parameters\end{tabular} 		&	 \textbf{no} 	&	 \begin{tabular}{@{}c@{}}additional \\ classifier\end{tabular}	&	 \begin{tabular}{@{}c@{}}range of \\ kernel params\end{tabular} 		&	 $k=5$ 	&	 $k=5$ 	\\
			\hline
			\begin{tabular}{@{}c@{}}Gradient \\ weighting $\eta$\end{tabular} 		&	 momentum 	&	 momentum 	&	 momentum 		&	 momentum 	&	 adadelta 	\\
			\hline
			\begin{tabular}{@{}c@{}}Learn.  rate\end{tabular} 		&	 $10^{-3}$ 	&	 $10^{-3}$ 	&	 cv 		&	 \textbf{default} 	&	 \textbf{no} 	\\
			\hline
			\begin{tabular}{@{}c@{}}Learn. rate \\ decay parameter\end{tabular} 		&	 \textbf{no} 	&	 yes 	&	 yes 		&	 $10^{-4}$	&	 \textbf{default} 	\\
			\hline
			\begin{tabular}{@{}c@{}}Data \\ augmentation\end{tabular} 		&	 yes 	&	 yes 	&	 yes 		&	 yes 	&	 \textbf{no} 	\\
			\hline
			\begin{tabular}{@{}c@{}}Weight  decay\end{tabular}		&	 yes 	&	 \textbf{no} 	&	 \textbf{no} 		&	 \textbf{no}  	&	 \textbf{no} 	\\
			\hline
		\end{tabular}
		\caption{Summary of parameter settings of state-of-the-art neural network approaches as applied on the Office dataset.}
		\label{tab:param_settings}
	\end{table}
	We repeat all evaluation five times with different random initializations and report the average accuracies and average ranks over all tasks in Table~\ref{tab:office_results}.
	\begin{table}[!ht]
		\renewcommand{\arraystretch}{1.3}
		\scriptsize
		\centering
		\begin{tabular}{|c||c|c|c|c|c|c||c|c|}
			\hline
			Method & A$\shortrightarrow$W & D$\shortrightarrow$W & W$\shortrightarrow$D & A$\shortrightarrow$D& D$\shortrightarrow$A& W$\shortrightarrow$A & Average & Average rank\\
			\hline\hline
			CNN~\cite{krizhevsky2012imagenet} & $52.9$ & $94.7$ & $99.0$ & $62.5$ & $50.2$ & $48.1$ & $67.9$ & $6.3$\\
			\hline
			TCA~\cite{pan2011domain} & $61.0$ & $95.4$ & $95.2$ & $60.8$ & $51.6$ & $50.9$ & $69.2$ & $6.0$\\
			\hline
			MMD~\cite{gretton2006kernel,long2015learning} & $63.8$ & $94.6$ & $98.8$ & $65.8$ & $52.8$ & $\mathit{51.9}$ & $71.3$ & $4.7$\\
			\hline
			CORAL~\cite{sun2016deep} & $\mathit{66.4}$ & $95.7$ & $99.2$ & $66.8$ & $52.8$ & $51.5$ & $72.1$ & $3.2$\\
			\hline
			DANN~\cite{ganin2016domain} & $\mathbf{73.0}$ & $\mathit{96.4}$ & $99.2$ & $\mathbf{72.3}$ & $53.4$ & $51.2$ & $\mathbf{74.3}$ & $\mathit{2.5}$\\
			\hline
			\hline
			CMD (ours) & $62.8$ & $\mathbf{96.7}$ & $\mathit{99.3}$ & $66.0$ & $\mathit{53.6}$ & $\mathit{51.9}$ & $71.7$ & $2.7$\\
			\hline
			FP-CMD (ours) & $64.8$ & $95.4$ & $\mathbf{99.4}$ & $\mathit{67.0}$ & $\mathbf{55.1}$ & $\mathbf{53.5}$ & $\mathit{72.5}$ & $\mathbf{2.0}$\\
			\hline
		\end{tabular}
		\caption{Classification accuracy on Office dataset for six domain adaptation scenarios (source$\shortrightarrow$target)}
		\label{tab:office_results}
	\end{table}
	
	Without considering the FP-CMD implementation, the CMD implementation shows the highest accuracy in four out of of six domain adaptation tasks.
	In the last two tasks, the DANN algorithm shows the highest accuracy and also has the highest average accuracy in these two scenarios.
	It is interesting to observe that this adversarial-based approach (DANN) outperforms others by more than $5\%$ classification accuracy.
	These two tasks have the Amazon (A) source domain in common containing only white background images.
	Consistently with other research works~\cite{tzeng2017adversarial} our experiments indicate that adversarial-based approaches work particularly well at aligning large domain shifts.
	However the instability issues of adversarial-based approaches identified in~\cite{arjovsky2017towards} can be the reason that our approach works better in the other domain adaptation tasks.
	
	The FP-CMD implementation shows the highest accuracy in three out of six tasks over all approaches and achieves the best average rank.
	In contrast to the other approaches, FP-CMD does so without data mirroring or rotation, without tuning, manually decreasing or cross-validating the learning rate, without applying different learning rates for different layers and without tuning of the domain adaptation weighting parameter $\lambda$ in Eq.\eqref{eq:obj} (see Table~\ref{tab:param_settings}).

	\subsection{Digit Recognition}
	\label{subsec:image_rec}
	
	In the following three domain adaptation experiments {SVHN$\rightarrow$MNIST}, {SynthDigits$\rightarrow$SVHN} and MNIST$\rightarrow$MNIST-M, we analyze the accuracy of our method based on the learning of deep convolutional neural networks on raw image data without using any additional knowledge.
	We used the provided training and test splits of the datasets described in Section~\ref{subsec:datasets}.
	
	In semi-supervised learning research it is often the case that the parameters of deep neural network architectures are specifically tuned for certain datasets~\cite{odena2018realistic}. This can cause problems when applying these methods to real-world applications.
	Since our goal is to propose a robust method, we rely on one architecture for all three digit recognition tasks.
	The architecture is not specifically developed for high performance of our method but rather independently developed in~\cite{haeusser2017associative}.
	In addition, we fix the learning rate, set the domain adaptation parameters to our default setting and change the activation function of the last layer to be the $\tanh$ function such that the output of the layer is bounded.
	
	We compare our algorithm to the following approaches:
	\begin{itemize}
		\item {\it Deep Convolutional Neural Network (CNN)}: The architecture of~\cite{haeusser2017associative} and trained via the Adam optimizer~\cite{kingma2015adam} as used by other methods~\cite{bousmalis2016domain,tzeng2017adversarial,sun2016deep}. Data augmentation is applied.
		
		\item {\it Deep Correlation Alignment (CORAL)}~\cite{sun2016deep}: The same optimization procedure and architecture as in CNN is used.
		The domain adaptation weighting parameter $\lambda$ is chosen in such a way that "at the end of training the classification loss and the CORAL loss are roughly the same"~\cite{sun2016deep}, i.e. $\lambda=1$ as in the original work.
		
		\item {\it Maximum Mean Discrepancy (MMD)}~\cite{gretton2006kernel}: We report the results of Bousmalis et al.~\cite{bousmalis2016domain} in which two separate architectures for each of the two tasks are trained by the Adam optimizer.
		The parameters are tuned according to the procedure reported in~\cite{long2015learning}.
		
		\item {\it Adversarial Discriminative Domain Adaptation (ADDA)~\cite{tzeng2017adversarial}}: We report results of the original paper for the SVHN$\rightarrow$MNIST task, based on the Adam optimizer.
		
		\item {\it Domain Adversarial Neural Networks (DANN)~\cite{ganin2016domain}}: The results of the original paper are reported. They used stochastic gradient descent with a polynomial decay rate, a momentum term and an exponential learning rate schedule.
		
		\item {\it Domain Separation Networks (DSN)~\cite{bousmalis2016domain}}: We report the results of the original work in which they used the adversarial approach as distance function for the similarity loss. Different architectures are used for both tasks. The hyper-parameters are tuned using a small labeled set from the target domain.
		
		\item {\it Central Moment Discrepancy (CMD)}: The approach of this paper with the same optimization strategy as in CNN, $k=5$ and $\lambda=1$ as described in Section~\ref{subsec:amazon_reviews}.
		
		\item {\it Cross-Variance Central Moment Discrepancy (CV-CMD)}: The approach of this paper including the alignment of all cross-variances, i.e. all monomials of order $2$ in Eq.\eqref{eq:nu}. The alignment term in the sum of the CMD is divided by $\sqrt{2}$ to compensate the higher number of second order terms. The parameters $k=5$ and $\lambda=1$ are used as in all other experiments.
	\end{itemize}
	
	\begin{table}[!ht]
		\renewcommand{\arraystretch}{1.3}
		\scriptsize
		\centering
		\begin{tabular}{|c||c|c|c||c|c|}
			\hline
			Method & $\xrightarrow[\text{MNIST}]{\text{SVHN}}$ & $\xrightarrow[\text{MNIST-M}]{\text{MNIST}}$ & $\xrightarrow[\text{SVHN}]{\text{SynthDigits}}$ & Average & Average rank\\
			\hline\hline
			CNN &  $66.74$ &	$70.85$ &	$80.94$ &	$72.84$ &	$7.3$\\
			\hline
			CORAL~\cite{sun2016deep} & $69.39$ & $77.34$ & $83.58$ & $76.77$ & $5.3$\\
			\hline
			ADDA~\cite{tzeng2017adversarial} & $76.00$ & $-$ &	$-$ &  	$76.00$ & $5.0$\\
			\hline
			MMD~\cite{gretton2006kernel,long2015learning} & $76.90$ & $71.10$ & $88.00$ & $78.67$ & $4.7$\\
			\hline
			DANN~\cite{ganin2016domain} & $76.66$ & $73.85$ & $\it 91.09$ & $80.53$ & $4.3$\\
			\hline
			DSN~\cite{bousmalis2016domain} &  $83.20$ & $82.70$ & $\bf 91.20$ & $\it 85.70$ & $\bf 2.3$\\
			\hline
			\hline
			CMD (ours) &  $\it 84.52$ & $\it 85.04$ & $85.52$ & $85.03$ & $\it 2.7$\\
			\hline
			CV-CMD (ours) &  $\bf 86.34$ & $\bf 88.03$ & $85.42$ & $\bf 86.60$ & $\bf2.3$\\
			\hline
		\end{tabular}
		\caption{Classification accuracy for three domain adaptation scenarios (source$\shortrightarrow$target) based on four large scale datasets~\cite{lecun1998mnist,netzer2011reading,ganin2016domain}.}
		\label{tab:digit_results}
	\end{table}
	
	The results are shown in Table~\ref{tab:digit_results}.
	Our method outperforms others in average accuracy as well as in average rank in the tasks SVHN$\rightarrow$MNIST and MNIST$\rightarrow$MNIST-M and performs worse on SynthDigits$\rightarrow$SVHN.
	
	At the SynthDigits$\rightarrow$SVHN task, the Proxy-$\mathcal{A}$ distance based (DANN, DSN) approaches perform better than distance based approaches without adversarial-based implementation (MMD, CORAL, CMD).
	Note that the performance gain (percentage over the baseline) of the best method on the SynthDigits$\rightarrow$SVHN task is rather low ($12.68\%$) compared to the other tasks ($29.37\%$ and $24.25\%$).
	That is, the methods perform more similar on this task than on the others \wrt this measure.
	
	The next section analyzes the classification accuracy sensitivity \wrt~changes of the hidden layer size and the number of moments.
	
	\subsection{Accuracy Sensitivity w.\,r.\,t.\, Parameter Changes}
	\label{subsection:sensitivity}
	
	The first sensitivity experiment aims at providing evidence regarding the sensitivity of the classification accuracy of the CMD regularizer \wrt~parameter changes of the parameter $k$.
	That is, the contribution of higher terms in the CMD are analyzed.
	The claim is that the accuracy of CMD-based networks does not depend strongly on the choice of $k$ in a range around its default value $5$.
	
\begin{figure}[ht]
		\centering
		\includegraphics[width=.75\textwidth]{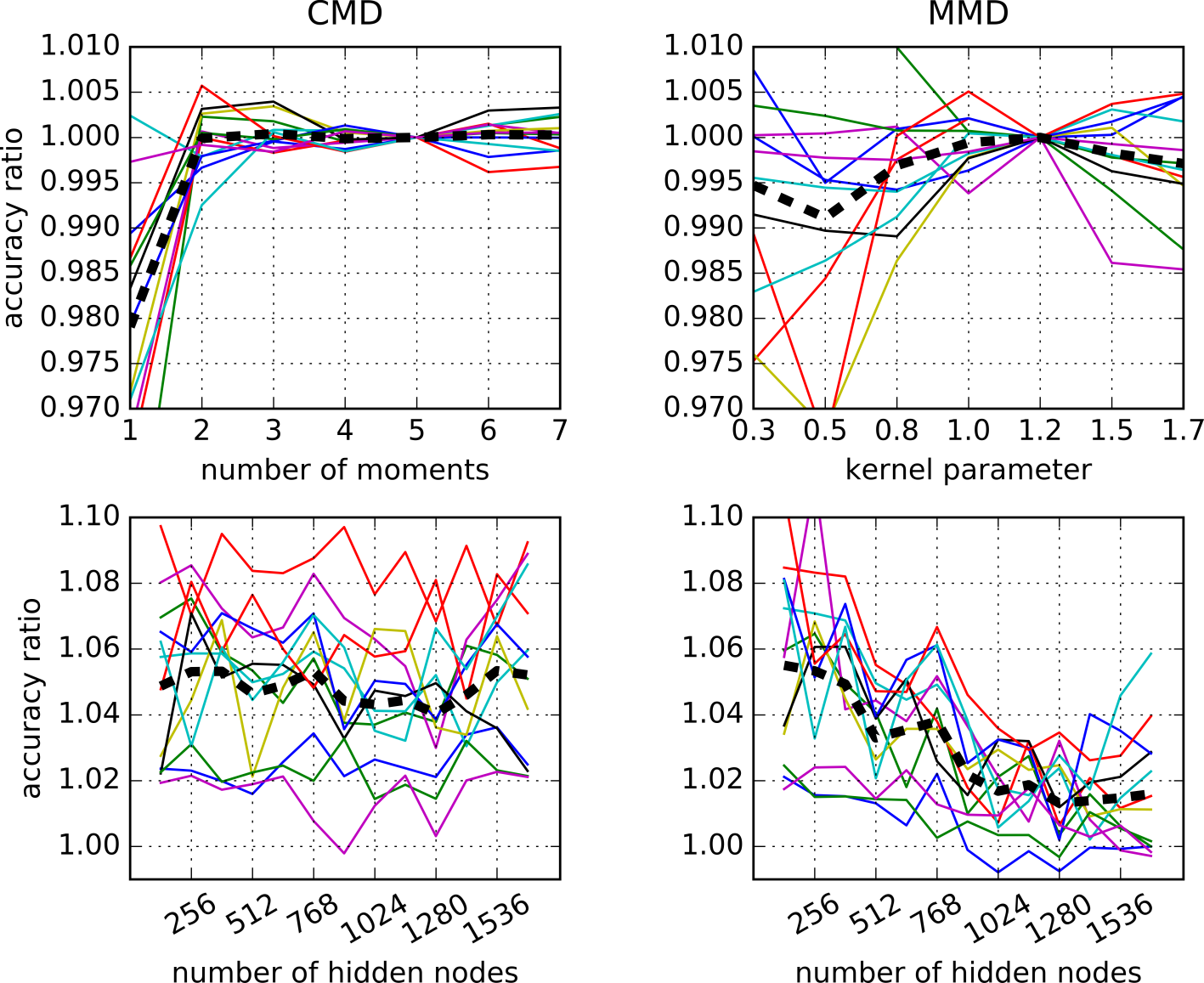}
		\caption[CMD accuracy sensitivity]{Sensitivity of classification accuracy \wrt different parameters of CMD (left) and MMD (right) on the Amazon reviews dataset.
			The horizontal axes show parameter values and the vertical axes show accuracy ratio values.
			Each line in the plots represents accuracy ratio values for one specific task.
			The ratio values in the upper left plot are computed \wrt the default accuracy for CMD ($k=5$) and on the right \wrt the best obtainable accuracy for MMD ($\beta=1.2$).
			The ratio values in the lower plots are computed \wrt the accuracies of the networks with the same hidden layer but without domain adaptation.}
		\label{fig:psens}
	\end{figure}

	In Fig.~\ref{fig:psens} we analyze the classification accuracy of a CMD-based network trained on all tasks of the Amazon reviews experiment.
	We perform a grid search for the parameter $k$ and the standard weighting parameter $\lambda$ in the objective in Eq.\eqref{eq:obj_approx}.
	We empirically choose a representative stable region for each parameter, $[0.3,3]$ for $\lambda$ and $\{1,\ldots,7\}$ for $k$.
	Since we want to analyze the sensitivity \wrt $k$, we averaged over the $\lambda$-dimension, resulting in one accuracy value per $k$ for each of the twelve tasks.
	Each accuracy is transformed into an accuracy ratio value by dividing it by the accuracy achieved with $k=5$.
	Thus, for each $k$ and each task, we calculate one value representing the ratio between the obtained accuracy and the accuracy of $k=5$.
	The results are shown in Fig.~\ref{fig:psens} in the upper left plot.
	The accuracy ratios between $k=5$ and $k\in\{3,4,6,7\}$ are lower than $0.5\%$, which underpins the claim that the accuracy of CMD-based networks does not depend strongly on the choice of $k$ in a range around its default value $5$.
	For $k=1$ and $k=2$ higher ratio values are obtained. In addition, for these two values many tasks show a worse accuracy than obtained by $k\in\{3,4,5,6,7\}$. From this, we additionally conclude that higher values of $k$ are preferable to $k=1$ and $k=2$.
	
	The same experimental procedure is performed with MMD regularization weighted by $\lambda\in[5,45]$ and Gaussian kernel parameter $\beta\in[0.3,1.7]$.
	We calculate the ratio values \wrt the accuracy of $\beta=1.2$, since this value of $\beta$ shows the highest mean accuracy of all tasks.
	Fig.~\ref{fig:psens} in the upper right plot shows the results.
	The accuracy of the MMD network is more sensitive to parameter changes than the CMD optimized version.
	Note that the problem of finding the best settings for parameter $\beta$ of the Gaussian kernel is a well known problem.
	
	The default number of hidden nodes in the sentiment analysis experiments in Section~\ref{subsec:amazon_reviews} is $50$ to be comparable with other state-of-the-art approaches~\cite{ganin2016domain}.
	The question arises whether the accuracy improvement of the CMD-regularization is robust to changes of the number of hidden nodes.
	
	In order to answer this question we calculate the accuracy ratio between the CMD-based network and the non-regularized network for each task of the Amazon reviews dataset for different numbers of hidden nodes in $\{128,256,384,\ldots,1664\}$.
	For higher numbers of hidden nodes our NN models do not converge with the optimization settings under consideration.
	For the parameters $\lambda$ and $k$ we use our default setting $\lambda=1$ and $k=5$.
	Fig.~\ref{fig:psens} in the lower left plot shows the ratio values (vertical axis) for every number of hidden nodes (horizontal axis) and every task (colored lines).
	The accuracy improvement of the CMD domain regularizer varies between $4\%$ and $6\%$.
	However, no significant accuracy ratio decrease can be observed.
	
	Fig.~\ref{fig:psens} shows that our default setting ($\lambda=1, k=5$) can be used independently of the number of hidden nodes for the sentiment analysis task.
	
	The same procedure is performed with the MMD weighted by parameter $\lambda=9$ and $\beta=1.2$ as these values show the highest classification accuracy for $50$ hidden nodes.
	Fig.~\ref{fig:psens} in the lower right plot shows that the accuracy improvement using the MMD decreases with increasing number of hidden nodes for this parameter setting.
	That is, for accurate performance of the MMD, additional parameter tuning procedures for $\lambda$ and $\beta$ need to be performed.
	
	\section{Proof details}
	\label{sec:proofs}
	
	\subsection{An Example of Mean Over-Penalization}
	\label{app:penalization_problem}
	
	Let the source distribution $\D_S$ be defined by the random variable $X_S=0.8\, Y+0.1$ with $Y$ following a Beta distribution with shape parameters $\alpha=\beta=0.4$ (Fig.~\ref{fig:problem} dashed). Let the left target distribution $\D_T^{(L)}$ be a Normal distribution with mean $0.5$ and variance $0.27^2$ (Fig.~\ref{fig:problem} left) and let the right target distribution $\D_T^{(R)}$ be defined by the random variable $X_T=0.8\cdot Y+0.12$ (Fig.~\ref{fig:problem} right). Then,
	\begin{align}
		d_{\Pk^1}(\D_S,\D_T^{(L)}) &= \big|\E_{\D_S}[x]-\E_{\D_T^{(L)}}[x]\big| \nonumber\\
		&= 0 < 0.02 < d_{\Pk^1}(\D_S,\D_T^{(R)}),
	\end{align}
	and for $\Pk^2$ and $\Pk^4$ it follows
	\begin{align}
		d_{\Pk^2}(\D_S,\D_T^{(L)})<0.016<0.02 < d_{\Pk^2}(\D_S,\D_T^{(R)})\\
		d_{\Pk^4}(\D_S,\D_T^{(L)})<0.02<0.021 < d_{\Pk^4}(\D_S,\D_T^{(R)}).
	\end{align}
	Let us now consider the MMD~\cite{gretton2006kernel} with standard polynomial kernel $\kappa_2(x,y)=(1+x y)^2$. It holds that
	\begin{align}
		\text{MMD}_{\kappa_2}&(\D_S,\D_T^{(L)})=\nonumber\\ =\,& \E_{\D_S}[\E_{\D_S}[\kappa(x,x')]]+\E_{\D_T^{(L)}}[\E_{\D_T^{(L)}}[\kappa(y,y')]]\nonumber\\
		& -2 \E_{\D_S}[\E_{\D_T^{(L)}}[\kappa(x,y)]]\nonumber\\
		=\,&2\,\big|\E_{\D_S}[x]-\E_{\D_T^{(L)}}[x]\big|^2+\big|\E_{\D_S}[x^2]-\E_{\D_T^{(L)}}[x^2]\big|^2\nonumber\\
		<\,& 0.00025 < 0.0012 < \text{MMD}_{\kappa_2}(\D_S,\D_T^{(R)}).
	\end{align}
	Similarly it follows for the quartic kernel $\kappa_4(x,y)=(1+x y)^4$ that
	\begin{align}
		\text{MMD}_{\kappa_4}(\D_S,\D_T^{(L)})<0.004<0.006<\text{MMD}_{\kappa_4}(\D_S,\D_T^{(R)}).
	\end{align}
	The mean and covariance feature matching integral probability metrics in~\cite{mroueh2017mcgan} coincide in our example with the integral probability metrics based on $\Pk^1$ and $\Pk^2$. Finally, for the CMD in Eq.\eqref{eq:cmd} with $a_1=\ldots=a_4=1$, we obtain
	\begin{align}
		\mathrm{cmd}_4(\D_S,\D_T^{(L)})>0.0207>0.02>\mathrm{cmd}_4(\D_S,\D_T^{(R)}).
	\end{align}
	
	\subsection{Proof of Theorem~\ref{thm:dual_cmd}}
	\label{app:dual}
	The proof follows from the linearity of the expectation for finite sums and the self-duality of the Euclidean norm.
	It holds that  
	\begin{align*}
		\label{eq:cmd_1}
		\mathrm{cmd}_k&(\D,\D') = \\
		=&\, a_1 \sup_{f\in\mathcal{P}^1}\big|\E_{\D}[f]-\E_{\D'}[f]\big| \nonumber \\
		&+\sum_{j=2}^k a_j \sup_{f\in\Pk^j}\big|\E_{\D}\left[f(\x-\E_{\D}[\x])\right] - \E_{\D'}\left[f(\x-\E_{\D'}[\x])\right]\big|\nonumber\\
		=&\, a_1 \sup_{\norm{\w}_2\leq 1}\big|\E_{\D}\left[\langle\w,\x\rangle_2\right] -\E_{\D'}\left[\langle\w,\x\rangle_2\right]\big| \nonumber\\
		&+ \sum_{j=2}^k a_j \sup_{\norm{\w}_2\leq 1}\big|\E_{\D}\big[\langle\w,\boldsymbol{\nu}^{(j)}(\x-\E_{\D}[\x])\rangle_2\big] \nonumber\\
		&\phantom{+ \sum_{j=2}^k \sup_{\norm{\w}_2\leq 1}\big|} -\E_{\D'}\big[\langle\w,\boldsymbol{\nu}^{(j)}(\x-\E_{\D'}[\x])\rangle_2\big]\big|\nonumber\\
		=&\, a_1 \sup_{\norm{\w}_2\leq 1} \big|\langle\w,\E_{\D}[\x]-\E_{\D'}[\x]\rangle_2\big|\\
		&+ \sum_{j=2}^k a_j \sup_{\norm{\w}_2\leq 1} \big|\big\langle\w, 	\E_{\D}[\boldsymbol{\nu}^{(j)}(\x-\E_{\D}[\x])] \\
		&\phantom{+ \sum_{j=2}^k \sup_{\norm{\w}_2\leq 1} \big|\langle\w,}
		-\E_{\D'}[\boldsymbol{\nu}^{(j)}(\x-\E_{\D'}[\x])]\big\rangle_2\big|\text{,}
	\end{align*}
	and finally $\mathrm{cmd}_k(\D,\D')  = \sum_{j=1}^k a_j \norm{c_j(\D)-c_j(\D')}_2$ $\blacksquare$
	
	\subsection{Proof of Theorem~\ref{thm:ch_fct}}
	\label{app:characteristic}
	We use the multi-index notations $\tv^{\boldsymbol{\alpha}}=t_1^{\alpha_1}\cdots t_m^{\alpha_m}$, ${\boldsymbol{\alpha}!}=\alpha_1!\cdots\alpha_m!$, $\boldsymbol{D}^{\boldsymbol{\alpha}}=D_1^{\alpha_1}\cdots D_m^{\alpha_m}$ and $|\boldsymbol{\alpha}|=\alpha_1+\ldots+\alpha_m$.
	
	Since all moments $c_{\boldsymbol{\alpha}}(\D)$ are finite, 
	the characteristic functions $\zeta_n,\zeta_\infty$ are analytic. 
	Note that 
	$c_{\boldsymbol{\alpha}}(\D_n) = (-i)^{|\boldsymbol{\alpha}|}\boldsymbol{D}^{\boldsymbol{\alpha}}\zeta_n(\tv)\big|_{\tv=\boldsymbol{0}}$ and therefore,
	\begin{align*}
		\zeta_n(\tv) &=\E_{\D_n} [e^{i\langle\tv,\x\rangle}] \\
		&= \sum_{|\boldsymbol{\alpha}|\geq 1} \frac{\tv^{\boldsymbol{\alpha}}}{\boldsymbol{\alpha}!}
		\E_{\D_n}[\boldsymbol{D}^{\boldsymbol{\alpha}}\zeta_n(\boldsymbol{0})]\\
		&= \sum_{|\boldsymbol{\alpha}|\geq 1} \frac{(-i)^{|\boldsymbol{\alpha}|} c_{\boldsymbol{\alpha}}(\D_n)}{\boldsymbol{\alpha}!} \tv^{\boldsymbol{\alpha}}\\
		&= \sum_{|\boldsymbol{\alpha}|\leq k} \frac{(-i)^{|\boldsymbol{\alpha}|} c_{\boldsymbol{\alpha}}(\D_n)}{\boldsymbol{\alpha}!} \tv^{\boldsymbol{\alpha}} 
		+
		\sum_{|\boldsymbol{\alpha}|= k+1} \frac{\tv^{\boldsymbol{\alpha}}}{\boldsymbol{\alpha}!} \boldsymbol{D}^{\boldsymbol{\alpha}} \zeta_n(\xi\cdot \tv)
	\end{align*}
	for some $\xi\in(0,1)$ by Taylor's formula with Lagrange's form of the remainder. 
	
	Let $k$ be odd, i.e., $|\boldsymbol{\alpha}|=k+1$ is even. Then, by integration and $|e^{i \langle\tv,\x \rangle}|\leq 1$ it follows 
	that 
	$\boldsymbol{D}^{\boldsymbol{\alpha}} \zeta_n(\xi\cdot \tv)\leq c_{\boldsymbol{\alpha}}(\D_n)$ and therefore,
	\begin{align*}
		|\zeta_n(\boldsymbol{t}) - \zeta_\infty(\boldsymbol{t})| \leq & 
		\sum_{|\boldsymbol{\alpha}|\leq k} \frac{|c_{\boldsymbol{\alpha}}(\D_n)-c_{\boldsymbol{\alpha}}(\D_\infty)|}{\boldsymbol{\alpha}!} 
		\tv^{\boldsymbol{\alpha}}\\
		&+ \sum_{|\boldsymbol{\alpha}|= k+1} \frac{\tv^{\boldsymbol{\alpha}}}{\boldsymbol{\alpha}!}
		\big(|c_{\boldsymbol{\alpha}}(\D_n)|+|c_{\boldsymbol{\alpha}}(\D_\infty)|\big)\\
		\leq & \sqrt{m}\cdot e^{\norm{\tv}_1}\cdot \mathrm{cmd}(\D_n,\D_\infty) \\
		&+ \frac{\norm{\tv}_1^{k+1}}{(k+1)!}\cdot \max_{|\boldsymbol{\alpha}|=k+1}\big(|c_{\boldsymbol{\alpha}}(\D_n)|+|c_{\boldsymbol{\alpha}}(\D_\infty)|\big)
	\end{align*}
	for all $\norm{\tv}_1\leq 1$.
	
	The characteristic functions $\zeta_n$ are analytic and thus, the uniform convergence on the unit interval implies pointwise convergence on $\mathbb{R}^m$. The weak convergence of the distributions follows from Levy's continuity theorem in multiple dimensions~\cite[Theorem 9.8.2]{dudley2002real}
	$\blacksquare$
	
	\subsection{Proof of Proposition~\ref{prop:bound}}
	\label{app:upper}
	Proposition~\ref{prop:bound} follows as the special case $m=1$ from the following more general proof.
	Let $c_j(\D)=\E_\D[\boldsymbol{\nu}^{(j)}(\vec x-\E_\D[\vec x])]$ be the central moment vector of $\D$ with $\boldsymbol{\nu}^{(j)}$ as defined in Eq.\eqref{eq:nu_marginal}.
	Then,
	\begin{align*}
		\frac{1}{|b-a|^j}&\|c_j(\D)-c_j(\D')\|_2 \leq\\
		&\leq 2 \sqrt{m} \max_{\D\in\mathcal{C}[a,b]} \left|\frac{c_j(\D)}{(b-a)^j}\right| \\
		& \leq  2 \sqrt{m} \max_{\D\in\mathcal{C}[a,b]} \E_{\D} \left[\left|\frac{x-\E_{\D}[x]}{b-a}\right|^j\right]
	\end{align*}
	where $\mathcal{C}[a,b]$ is the set of all one-dimensional $[a,b]$-supported distributions.
	By applying the Edmundson\--Mandansky inequality~\cite{madansky1959bounds} to the convex function $|(x-\E_{\D}[x])/(b-a)|^j$ and symmetry arguments as in~\cite{egozcue2012smallest}, we get
	\begin{align*}
		\E_{\D} \left[\left|\frac{x-\E_{\D}[x]}{b-a}\right|^j\right] \leq &\frac{b-\E_\D[ x]}{b-a} \cdot\left|\frac{a-\E_{\D}[x]}{b-a}\right|^j\\
		\phantom{\leq} &+ \frac{\E_{\D}[x]-a}{b-a} \cdot \left|\frac{b-\E_{\D}[x]}{b-a}\right|^j\\
		\leq &\max_{x\in[0,1]}\left( (1-x) x^j + (1-x)^j x\right)\\
		= &\max_{x\in[0,1/2]}\left( (1-x) x^j + (1-x)^j x\right)\\
		\leq &\max_{x\in[0,1/2]} (1-x) x^j + \max_{x\in[0,1/2]} (1-x)^j x\\
		\leq &\frac{1}{j+1}\left(\frac{j}{j+1}\right)^j
		+ \frac{1}{2^{1+j}}\,\blacksquare
	\end{align*}
	
	\subsection{Derivation of Gradients}
	\label{app:gradients}
	Here, we derive the gradients of the CMD estimate in Eq.\eqref{eq:cmd_estimate} for the neural network architecture in 
	Section~\ref{sec:MANN}.
	Let the mean $\E [X]$ of the sample $X$ be defined by 
	$\E[X] = \frac{1}{|X|}\sum_{\x\in X} \x$ and the sampled central moments ${\E[\boldsymbol{\nu}^{(k)}(X-\E[X])]}$, 
	with the set notations 
	\begin{align}
		X-\E[X] &:= \{\x-\E[X] | \x\in X\},\\
		\boldsymbol{\nu}^{(k)}(X) &:= \{\boldsymbol{\nu}^{(k)}(\x)|\x\in X\}. 
	\end{align}
	
	Let $\odot$ be the coordinate-wise multiplication. Then, by setting
	\begin{align}
		\nabla_{\vec b}~\mathrm{cmd} &:= \nabla_{\vec b}~\mathrm{cmd}(h_0(X_S),h_0(X_T)), \\
		\nabla_{\vec W}~\mathrm{cmd} &:= \nabla_{\vec W}~\mathrm{cmd}(h_0(X_S),h_0(X_T)), \\
		\boldsymbol{\Gamma}_{j,X} &:=  \boldsymbol{\nu}^{(j)}(h_0(X)-\E[h_0(X)]),\\
		\boldsymbol{\Delta}_{X_S,X_T} &:= h_0(X_S)- h_0(X_T), \\
		\boldsymbol{q}_X &:= h_0(X)\odot (\vec 1 - h_0(X)),
	\end{align}
	the application of the chain rule gives
	\begin{align}
		\nabla_{\vec b}~\mathrm{cmd} 
		= &\nabla_{\vec b} \norm{\E[\boldsymbol{\Delta}_{X_S,X_T}]}_2
		+\sum_{j=2}^{k} \nabla_{\vec b} \| \E[\boldsymbol{\Gamma}_{j,X_S}] -  \E[\boldsymbol{\Gamma}_{j,X_T}]\|_2 \nonumber\\
		= &\frac{\E[\boldsymbol{\Delta}_{X_S,X_T}]\odot 
			(\E[\boldsymbol{q}_{X_S}] - \E[\boldsymbol{q}_{X_T}])}{\norm{\E[\boldsymbol{\Delta}_{X_S,X_T}]}_2}
		+\sum_{j=2}^{k} \frac{{\E} [\boldsymbol{\Gamma}_{j,X_S}] -{\E}[\boldsymbol{\Gamma}_{j,X_T}]}
		{\norm{\E[\boldsymbol{\Gamma}_{j,X_S}] -\E[\boldsymbol{\Gamma}_{j,X_T}]}_2} \nonumber\\
		&\phantom{= + \sum_{j=2}^{k}}\odot (\E[\nabla_{\vec b} 
		\boldsymbol{\Gamma}_{j,X_S}]  - \E[\nabla_{\vec b}
		\boldsymbol{\Gamma}_{j,X_T}])
	\end{align}
	and
	\begin{align}
		\nabla_{\vec b} \boldsymbol{\Gamma}_{j,X} = j\cdot \boldsymbol{\Gamma}_{j-1,X} 
		\odot 
		(\boldsymbol{q}_X - \E[\boldsymbol{q}_X]),
	\end{align}
	which follows from  
	\begin{align}
		\nabla_{\x}~\mathrm{sigm}(\x)=\mathrm{sigm}(\x)\odot (1-\mathrm{sigm}(\x)).
	\end{align}
	Analogously, we obtain $\nabla_{\vec W}~\mathrm{cmd}$.
	The gradients of the cross-entropy loss function \wrt $\vec W$, $\vec b$, $\vec V$ and $\vec c$ are
	\begin{align}
		\nabla_{\vec c}~\mathcal{L}(h(X_S),Y_S) =\, &\E[h_1(X_S)- Y_S],\\
		\nabla_{\vec V}~\mathcal{L}(h(X_S),Y_S) =\, &\E[(h_1(X_S)- Y_S)\cdot h_1(X_S)^T], \\
		\nabla_{\vec b}~\mathcal{L}(h(X_S),Y_S) =\, &\E[\vec V^T (h_1(X_S)- Y_S)\nonumber \\
		&\phantom{\E[}\odot h_1(X) \odot (\vec 1 - h_1(X))],\\
		\nabla_{\vec W}~\mathcal{L}(h(X_S),Y_S) =\, &\E[(\vec V^T (h_1(X_S)- Y_S) \odot h_1(X)\nonumber\\
		&\phantom{\E[} \odot (\vec 1 - h_1(X))) \cdot X_S^T].
	\end{align}
	
	\section{Conclusion}
	\label{sec:conclusion}
	
	We proposed a novel approach for unsupervised domain-adaptation for neural networks that relies on a metric-based regularization of the learning process.
	The regularization aims at maximizing the similarity of domain-specific activation distributions by minimizing the proposed Central Moment Discrepancy (CMD) metric.
	The CMD solves instability issues that arise in the application of integral probability metrics based on polynomial function spaces.
	We  proved further theoretically properties of the CMD including a relation to weak convergence of distributions, a strictly decreasing upper bound for its moment terms and a computationally efficient dual representation.
	We empirically analyzed the classification performance of the CMD on an artificial dataset and $21$ standard benchmark tasks for domain adaptation based on $6$ datasets.
	The proposed approach is robust \wrt theoretical and practical aspects while it shows higher classification accuracies than comparable state-of-the-art approaches on most domain adaptation tasks.
	
	In this work, we used a sub-optimal fixed default parameter setting for all experiments.
	It is part of future work to develop an unsupervised model selection method that enables further accuracy improvement.
	Another open question is how to extend the current approach to multiple domains.
	Improved theoretical target error bounds are also part of our future work.
	
	\section*{Acknowledgements}
	This work was partially funded by SCCH within the Austrian COMET programme.
	We thank Florian Sobieczky and Ramin Nikzad-Langerodi for helpful discussions.
	
	\section*{References}
	
	\bibliographystyle{model5-names}
	\bibliography{cmd_journal}

\begin{thebibliography}{50}
\expandafter\ifx\csname natexlab\endcsname\relax\def\natexlab#1{#1}\fi
\providecommand{\url}[1]{\texttt{#1}}
\providecommand{\href}[2]{#2}
\providecommand{\path}[1]{#1}
\providecommand{\DOIprefix}{doi:}
\providecommand{\ArXivprefix}{arXiv:}
\providecommand{\URLprefix}{URL: }
\providecommand{\Pubmedprefix}{pmid:}
\providecommand{\doi}[1]{\href{http://dx.doi.org/#1}{\path{#1}}}
\providecommand{\Pubmed}[1]{\href{pmid:#1}{\path{#1}}}
\providecommand{\bibinfo}[2]{#2}
\ifx\xfnm\relax \def\xfnm[#1]{\unskip,\space#1}\fi
\bibitem[{Arjovsky \& Bottou(2017)}]{arjovsky2017towards}
\bibinfo{author}{Arjovsky, M.}, \& \bibinfo{author}{Bottou, L.}
  (\bibinfo{year}{2017}).
\newblock \bibinfo{title}{Towards principled methods for training generative
  adversarial networks}.
\newblock In {\it \bibinfo{booktitle}{International Conference on Learning
  Representations}\/}.
\bibitem[{Baktashmotlagh et~al.(2013)Baktashmotlagh, Harandi, Lovell \&
  Salzmann}]{baktashmotlagh2013unsupervised}
\bibinfo{author}{Baktashmotlagh, M.}, \bibinfo{author}{Harandi, M.~T.},
  \bibinfo{author}{Lovell, B.~C.}, \& \bibinfo{author}{Salzmann, M.}
  (\bibinfo{year}{2013}).
\newblock \bibinfo{title}{Unsupervised domain adaptation by domain invariant
  projection}.
\newblock In {\it \bibinfo{booktitle}{IEEE International Conference on Computer
  Vision}\/} (pp. \bibinfo{pages}{769--776}).
\bibitem[{Ben-David et~al.(2010)Ben-David, Blitzer, Crammer, Kulesza, Pereira
  \& Vaughan}]{ben2010theory}
\bibinfo{author}{Ben-David, S.}, \bibinfo{author}{Blitzer, J.},
  \bibinfo{author}{Crammer, K.}, \bibinfo{author}{Kulesza, A.},
  \bibinfo{author}{Pereira, F.}, \& \bibinfo{author}{Vaughan, J.~W.}
  (\bibinfo{year}{2010}).
\newblock \bibinfo{title}{A theory of learning from different domains}.
\newblock {\it \bibinfo{journal}{Machine learning}\/},  {\it
  \bibinfo{volume}{79}\/}, \bibinfo{pages}{151--175}.
\bibitem[{Blitzer et~al.(2006)Blitzer, McDonald \& Pereira}]{blitzer2006domain}
\bibinfo{author}{Blitzer, J.}, \bibinfo{author}{McDonald, R.}, \&
  \bibinfo{author}{Pereira, F.} (\bibinfo{year}{2006}).
\newblock \bibinfo{title}{Domain adaptation with structural correspondence
  learning}.
\newblock In {\it \bibinfo{booktitle}{Conference on Empirical Methods in
  Natural Language Processing}\/} (pp. \bibinfo{pages}{120--128}).
\newblock \bibinfo{organization}{Association for Computational Linguistics}.
\bibitem[{Bousmalis et~al.(2016)Bousmalis, Trigeorgis, Silberman, Krishnan \&
  Erhan}]{bousmalis2016domain}
\bibinfo{author}{Bousmalis, K.}, \bibinfo{author}{Trigeorgis, G.},
  \bibinfo{author}{Silberman, N.}, \bibinfo{author}{Krishnan, D.}, \&
  \bibinfo{author}{Erhan, D.} (\bibinfo{year}{2016}).
\newblock \bibinfo{title}{Domain separation networks}.
\newblock In {\it \bibinfo{booktitle}{Advances in Neural Information Processing
  Systems}\/} (pp. \bibinfo{pages}{343--351}).
\bibitem[{Chen et~al.(2012)Chen, Xu, Weinberger \& Sha}]{chen2012marginalized}
\bibinfo{author}{Chen, M.}, \bibinfo{author}{Xu, Z.},
  \bibinfo{author}{Weinberger, K.}, \& \bibinfo{author}{Sha, F.}
  (\bibinfo{year}{2012}).
\newblock \bibinfo{title}{Marginalized denoising autoencoders for domain
  adaptation}.
\newblock In {\it \bibinfo{booktitle}{International Conference on Machine
  Learning}\/} (pp. \bibinfo{pages}{767--774}).
\bibitem[{Chollet(2015)}]{chollet2015keras}
\bibinfo{author}{Chollet, F.} (\bibinfo{year}{2015}).
\newblock \bibinfo{title}{Keras: Deep learning library for theano and
  tensorflow}.
\bibitem[{Cortes \& Mohri(2014)}]{cortes2014domain}
\bibinfo{author}{Cortes, C.}, \& \bibinfo{author}{Mohri, M.}
  (\bibinfo{year}{2014}).
\newblock \bibinfo{title}{Domain adaptation and sample bias correction theory
  and algorithm for regression}.
\newblock {\it \bibinfo{journal}{Theoretical Computer Science}\/},  {\it
  \bibinfo{volume}{519}\/}, \bibinfo{pages}{103--126}.
\bibitem[{Duchi et~al.(2011)Duchi, Hazan \& Singer}]{duchi2011adaptive}
\bibinfo{author}{Duchi, J.}, \bibinfo{author}{Hazan, E.}, \&
  \bibinfo{author}{Singer, Y.} (\bibinfo{year}{2011}).
\newblock \bibinfo{title}{Adaptive subgradient methods for online learning and
  stochastic optimization}.
\newblock {\it \bibinfo{journal}{Journal of Machine Learning Research}\/},
  {\it \bibinfo{volume}{12}\/}, \bibinfo{pages}{2121--2159}.
\bibitem[{Dudley(2002)}]{dudley2002real}
\bibinfo{author}{Dudley, R.~M.} (\bibinfo{year}{2002}).
\newblock {\it \bibinfo{title}{Real analysis and probability}\/}
  volume~\bibinfo{volume}{74}.
\newblock \bibinfo{publisher}{Cambridge University Press}.
\bibitem[{Egozcue et~al.(2012)Egozcue, Garc{\'\i}a, Wong \&
  Zitikis}]{egozcue2012smallest}
\bibinfo{author}{Egozcue, M.}, \bibinfo{author}{Garc{\'\i}a, L.~F.},
  \bibinfo{author}{Wong, W.-K.}, \& \bibinfo{author}{Zitikis, R.}
  (\bibinfo{year}{2012}).
\newblock \bibinfo{title}{The smallest upper bound for the pth absolute central
  moment of a class of random variables}.
\newblock {\it \bibinfo{journal}{Mathematical Scientist}\/},  {\it
  \bibinfo{volume}{37}\/}.
\bibitem[{Fukumizu et~al.(2009)Fukumizu, Gretton, Lanckriet, Sch{\"o}lkopf \&
  Sriperumbudur}]{fukumizu2009kernel}
\bibinfo{author}{Fukumizu, K.}, \bibinfo{author}{Gretton, A.},
  \bibinfo{author}{Lanckriet, G.~R.}, \bibinfo{author}{Sch{\"o}lkopf, B.}, \&
  \bibinfo{author}{Sriperumbudur, B.~K.} (\bibinfo{year}{2009}).
\newblock \bibinfo{title}{Kernel choice and classifiability for {RKHS}
  embeddings of probability distributions}.
\newblock In {\it \bibinfo{booktitle}{Advances in Neural Information Processing
  Systems}\/} (pp. \bibinfo{pages}{1750--1758}).
\bibitem[{Ganin et~al.(2016)Ganin, Ustinova, Ajakan, Germain, Larochelle,
  Laviolette, Marchand \& Lempitsky}]{ganin2016domain}
\bibinfo{author}{Ganin, Y.}, \bibinfo{author}{Ustinova, E.},
  \bibinfo{author}{Ajakan, H.}, \bibinfo{author}{Germain, P.},
  \bibinfo{author}{Larochelle, H.}, \bibinfo{author}{Laviolette, F.},
  \bibinfo{author}{Marchand, M.}, \& \bibinfo{author}{Lempitsky, V.}
  (\bibinfo{year}{2016}).
\newblock \bibinfo{title}{Domain-adversarial training of neural networks}.
\newblock {\it \bibinfo{journal}{Journal of Machine Learning Research}\/},
  (pp. \bibinfo{pages}{2962--2971}).
\bibitem[{Glorot et~al.(2011)Glorot, Bordes \& Bengio}]{glorot2011domain}
\bibinfo{author}{Glorot, X.}, \bibinfo{author}{Bordes, A.}, \&
  \bibinfo{author}{Bengio, Y.} (\bibinfo{year}{2011}).
\newblock \bibinfo{title}{Domain adaptation for large-scale sentiment
  classification: A deep learning approach}.
\newblock In {\it \bibinfo{booktitle}{International Conference on Machine
  Learning}\/} (pp. \bibinfo{pages}{513--520}).
\bibitem[{Gretton et~al.(2006)Gretton, Borgwardt, Rasch, Sch{\"o}lkopf \&
  Smola}]{gretton2006kernel}
\bibinfo{author}{Gretton, A.}, \bibinfo{author}{Borgwardt, K.~M.},
  \bibinfo{author}{Rasch, M.}, \bibinfo{author}{Sch{\"o}lkopf, B.}, \&
  \bibinfo{author}{Smola, A.~J.} (\bibinfo{year}{2006}).
\newblock \bibinfo{title}{A kernel method for the two-sample-problem}.
\newblock In {\it \bibinfo{booktitle}{Advances in Neural Information Processing
  Systems}\/} (pp. \bibinfo{pages}{513--520}).
\bibitem[{Gretton et~al.(2012)Gretton, Sejdinovic, Strathmann, Balakrishnan,
  Pontil, Fukumizu \& Sriperumbudur}]{gretton2012optimal}
\bibinfo{author}{Gretton, A.}, \bibinfo{author}{Sejdinovic, D.},
  \bibinfo{author}{Strathmann, H.}, \bibinfo{author}{Balakrishnan, S.},
  \bibinfo{author}{Pontil, M.}, \bibinfo{author}{Fukumizu, K.}, \&
  \bibinfo{author}{Sriperumbudur, B.~K.} (\bibinfo{year}{2012}).
\newblock \bibinfo{title}{Optimal kernel choice for large-scale two-sample
  tests}.
\newblock In {\it \bibinfo{booktitle}{Advances in Neural Information Processing
  Systems}\/} (pp. \bibinfo{pages}{1205--1213}).
\bibitem[{Haeusser et~al.(2017)Haeusser, Frerix, Mordvintsev \&
  Cremers}]{haeusser2017associative}
\bibinfo{author}{Haeusser, P.}, \bibinfo{author}{Frerix, T.},
  \bibinfo{author}{Mordvintsev, A.}, \& \bibinfo{author}{Cremers, D.}
  (\bibinfo{year}{2017}).
\newblock \bibinfo{title}{Associative domain adaptation}.
\newblock In {\it \bibinfo{booktitle}{International Conference on Computer
  Vision (ICCV)}\/} (pp. \bibinfo{pages}{2784--2792}).
\bibitem[{Jacobsen et~al.(2018)Jacobsen, Smeulders \&
  Oyallon}]{jacobsen2018revnet}
\bibinfo{author}{Jacobsen, J.-H.}, \bibinfo{author}{Smeulders, A.}, \&
  \bibinfo{author}{Oyallon, E.} (\bibinfo{year}{2018}).
\newblock \bibinfo{title}{i-revnet: Deep invertible networks}.
\newblock In {\it \bibinfo{booktitle}{International Conference on Learning
  Representations}\/}.
\bibitem[{Kingma \& Ba(2015)}]{kingma2015adam}
\bibinfo{author}{Kingma, D.~P.}, \& \bibinfo{author}{Ba, J.}
  (\bibinfo{year}{2015}).
\newblock \bibinfo{title}{Adam: A method for stochastic optimization}.
\newblock In {\it \bibinfo{booktitle}{International Conference on Learning
  Representations}\/}.
\bibitem[{Krizhevsky et~al.(2012)Krizhevsky, Sutskever \&
  Hinton}]{krizhevsky2012imagenet}
\bibinfo{author}{Krizhevsky, A.}, \bibinfo{author}{Sutskever, I.}, \&
  \bibinfo{author}{Hinton, G.~E.} (\bibinfo{year}{2012}).
\newblock \bibinfo{title}{Imagenet classification with deep convolutional
  neural networks}.
\newblock In {\it \bibinfo{booktitle}{Advances in Neural Information Processing
  Systems}\/} (pp. \bibinfo{pages}{1097--1105}).
\bibitem[{LeCun(1998)}]{lecun1998mnist}
\bibinfo{author}{LeCun, Y.} (\bibinfo{year}{1998}).
\newblock \bibinfo{title}{The mnist database of handwritten digits}.
\newblock \bibinfo{note}{Http://yann.lecun.com/exdb/mnist/}.
\bibitem[{LeCun et~al.(2015)LeCun, Bengio \& Hinton}]{lecun2015deep}
\bibinfo{author}{LeCun, Y.}, \bibinfo{author}{Bengio, Y.}, \&
  \bibinfo{author}{Hinton, G.} (\bibinfo{year}{2015}).
\newblock \bibinfo{title}{Deep learning}.
\newblock {\it \bibinfo{journal}{Nature}\/},  {\it \bibinfo{volume}{521}\/},
  \bibinfo{pages}{436--444}.
\bibitem[{Li et~al.(2017{\natexlab{a}})Li, Chang, Cheng, Yang \&
  P{\'o}czos}]{li2017mmd}
\bibinfo{author}{Li, C.-L.}, \bibinfo{author}{Chang, W.-C.},
  \bibinfo{author}{Cheng, Y.}, \bibinfo{author}{Yang, Y.}, \&
  \bibinfo{author}{P{\'o}czos, B.} (\bibinfo{year}{2017}{\natexlab{a}}).
\newblock \bibinfo{title}{{MMD} gan: Towards deeper understanding of moment
  matching network}.
\newblock In {\it \bibinfo{booktitle}{Advances in Neural Information Processing
  Systems}\/} (pp. \bibinfo{pages}{2203--2213}).
\bibitem[{Li et~al.(2017{\natexlab{b}})Li, Song \& Huang}]{li2017prediction}
\bibinfo{author}{Li, S.}, \bibinfo{author}{Song, S.}, \&
  \bibinfo{author}{Huang, G.} (\bibinfo{year}{2017}{\natexlab{b}}).
\newblock \bibinfo{title}{Prediction reweighting for domain adaptation}.
\newblock {\it \bibinfo{journal}{IEEE Transactions on Neural Networks and
  Learning Systems}\/},  {\it \bibinfo{volume}{28}\/},
  \bibinfo{pages}{1682--1695}.
\bibitem[{Li et~al.(2017{\natexlab{c}})Li, Wang, Shi, Liu \&
  Hou}]{li2016revisiting}
\bibinfo{author}{Li, Y.}, \bibinfo{author}{Wang, N.}, \bibinfo{author}{Shi,
  J.}, \bibinfo{author}{Liu, J.}, \& \bibinfo{author}{Hou, X.}
  (\bibinfo{year}{2017}{\natexlab{c}}).
\newblock \bibinfo{title}{Revisiting batch normalization for practical domain
  adaptation}.
\newblock In {\it \bibinfo{booktitle}{International Conference on Learning
  Representations Workshop}\/}.
\bibitem[{Long et~al.(2015)Long, Cao, Wang \& Jordan}]{long2015learning}
\bibinfo{author}{Long, M.}, \bibinfo{author}{Cao, Y.}, \bibinfo{author}{Wang,
  J.}, \& \bibinfo{author}{Jordan, M.} (\bibinfo{year}{2015}).
\newblock \bibinfo{title}{Learning transferable features with deep adaptation
  networks}.
\newblock In {\it \bibinfo{booktitle}{Proceedings of the International
  Conference on Machine Learning}\/} (pp. \bibinfo{pages}{97--105}).
\bibitem[{Long et~al.(2017)Long, Wang \& Jordan}]{long2016joint}
\bibinfo{author}{Long, M.}, \bibinfo{author}{Wang, J.}, \&
  \bibinfo{author}{Jordan, M.~I.} (\bibinfo{year}{2017}).
\newblock \bibinfo{title}{Deep transfer learning with joint adaptation
  networks}.
\newblock In {\it \bibinfo{booktitle}{International Conference on Machine
  Learning}\/} (pp. \bibinfo{pages}{2208--2217}).
\bibitem[{Long et~al.(2016)Long, Zhu, Wang \& Jordan}]{long2016unsupervised}
\bibinfo{author}{Long, M.}, \bibinfo{author}{Zhu, H.}, \bibinfo{author}{Wang,
  J.}, \& \bibinfo{author}{Jordan, M.~I.} (\bibinfo{year}{2016}).
\newblock \bibinfo{title}{Unsupervised domain adaptation with residual transfer
  networks}.
\newblock In {\it \bibinfo{booktitle}{Advances in Neural Information Processing
  Systems}\/} (pp. \bibinfo{pages}{136--144}).
\bibitem[{Louizos et~al.(2016)Louizos, Swersky, Li, Welling \&
  Zemel}]{louizos2015variational}
\bibinfo{author}{Louizos, C.}, \bibinfo{author}{Swersky, K.},
  \bibinfo{author}{Li, Y.}, \bibinfo{author}{Welling, M.}, \&
  \bibinfo{author}{Zemel, R.} (\bibinfo{year}{2016}).
\newblock \bibinfo{title}{The variational fair auto encoder}.
\newblock In {\it \bibinfo{booktitle}{International Conference on Learning
  Representations}\/}.
\bibitem[{Madansky(1959)}]{madansky1959bounds}
\bibinfo{author}{Madansky, A.} (\bibinfo{year}{1959}).
\newblock \bibinfo{title}{Bounds on the expectation of a convex function of a
  multivariate random variable}.
\newblock {\it \bibinfo{journal}{The Annals of Mathematical Statistics}\/},
  {\it \bibinfo{volume}{30}\/}, \bibinfo{pages}{743--746}.
\bibitem[{Mansour et~al.(2009)Mansour, Mohri \&
  Rostamizadeh}]{mansour2009domain}
\bibinfo{author}{Mansour, Y.}, \bibinfo{author}{Mohri, M.}, \&
  \bibinfo{author}{Rostamizadeh, A.} (\bibinfo{year}{2009}).
\newblock \bibinfo{title}{Domain adaptation: Learning bounds and algorithms}.
\newblock In {\it \bibinfo{booktitle}{Conference on Learning Theory}\/}.
\bibitem[{Mroueh et~al.(2017)Mroueh, Sercu \& Goel}]{mroueh2017mcgan}
\bibinfo{author}{Mroueh, Y.}, \bibinfo{author}{Sercu, T.}, \&
  \bibinfo{author}{Goel, V.} (\bibinfo{year}{2017}).
\newblock \bibinfo{title}{{McGAN}: Mean and covariance feature matching gan}.
\newblock In {\it \bibinfo{booktitle}{International Conference on Machine
  Learning}\/} (pp. \bibinfo{pages}{2527--2535}).
\bibitem[{M{\"u}ller(1997)}]{muller1997integral}
\bibinfo{author}{M{\"u}ller, A.} (\bibinfo{year}{1997}).
\newblock \bibinfo{title}{Integral probability metrics and their generating
  classes of functions}.
\newblock {\it \bibinfo{journal}{Advances in Applied Probability}\/},  {\it
  \bibinfo{volume}{29}\/}, \bibinfo{pages}{429--443}.
\bibitem[{Netzer et~al.(2011)Netzer, Wang, Coates, Bissacco, Wu \&
  Ng}]{netzer2011reading}
\bibinfo{author}{Netzer, Y.}, \bibinfo{author}{Wang, T.},
  \bibinfo{author}{Coates, A.}, \bibinfo{author}{Bissacco, A.},
  \bibinfo{author}{Wu, B.}, \& \bibinfo{author}{Ng, A.~Y.}
  (\bibinfo{year}{2011}).
\newblock \bibinfo{title}{Reading digits in natural images with unsupervised
  feature learning}.
\newblock In {\it \bibinfo{booktitle}{Advances in Neural Information Processing
  Systems Workshop on Deep Learning and Unsupervised Feature Learning}\/}.
\bibitem[{Nikzad-Langerodi et~al.(2018)Nikzad-Langerodi, Zellinger, Lughofer \&
  Saminger-Platz}]{nikzad2018nipals}
\bibinfo{author}{Nikzad-Langerodi, R.}, \bibinfo{author}{Zellinger, W.},
  \bibinfo{author}{Lughofer, E.}, \& \bibinfo{author}{Saminger-Platz, S.}
  (\bibinfo{year}{2018}).
\newblock \bibinfo{title}{Domain-invariant partial-least-squares regression}.
\newblock {\it \bibinfo{journal}{Analytical chemistry}\/},  {\it
  \bibinfo{volume}{90}\/}, \bibinfo{pages}{6693--6701}.
\bibitem[{Odena et~al.(2018)Odena, Oliver, Raffel, Cubuk \&
  Goodfellow}]{odena2018realistic}
\bibinfo{author}{Odena, A.}, \bibinfo{author}{Oliver, A.},
  \bibinfo{author}{Raffel, C.}, \bibinfo{author}{Cubuk, E.~D.}, \&
  \bibinfo{author}{Goodfellow, I.} (\bibinfo{year}{2018}).
\newblock \bibinfo{title}{Realistic evaluation of semi-supervised learning
  algorithms}.
\newblock In {\it \bibinfo{booktitle}{International Conference on Learning
  Representations Workshop}\/}.
\bibitem[{Pan et~al.(2011)Pan, Tsang, Kwok \& Yang}]{pan2011domain}
\bibinfo{author}{Pan, S.~J.}, \bibinfo{author}{Tsang, I.~W.},
  \bibinfo{author}{Kwok, J.~T.}, \& \bibinfo{author}{Yang, Q.}
  (\bibinfo{year}{2011}).
\newblock \bibinfo{title}{Domain adaptation via transfer component analysis}.
\newblock {\it \bibinfo{journal}{IEEE Transactions on Neural Networks}\/},
  {\it \bibinfo{volume}{22}\/}, \bibinfo{pages}{199--210}.
\bibitem[{Pan \& Yang(2010)}]{pan2010survey}
\bibinfo{author}{Pan, S.~J.}, \& \bibinfo{author}{Yang, Q.}
  (\bibinfo{year}{2010}).
\newblock \bibinfo{title}{A survey on transfer learning}.
\newblock {\it \bibinfo{journal}{IEEE Transactions on Knowledge and Data
  Engineering}\/},  {\it \bibinfo{volume}{22}\/}, \bibinfo{pages}{1345--1359}.
\bibitem[{Rachev et~al.(2013)Rachev, Klebanov, Stoyanov \&
  Fabozzi}]{rachev2013methods}
\bibinfo{author}{Rachev, S.~T.}, \bibinfo{author}{Klebanov, L.},
  \bibinfo{author}{Stoyanov, S.~V.}, \& \bibinfo{author}{Fabozzi, F.}
  (\bibinfo{year}{2013}).
\newblock {\it \bibinfo{title}{The methods of distances in the theory of
  probability and statistics}\/}.
\newblock \bibinfo{publisher}{Springer Science \& Business Media}.
\bibitem[{Rui et~al.(1999)Rui, Huang \& Chang}]{rui1999image}
\bibinfo{author}{Rui, Y.}, \bibinfo{author}{Huang, T.~S.}, \&
  \bibinfo{author}{Chang, S.-F.} (\bibinfo{year}{1999}).
\newblock \bibinfo{title}{Image retrieval: Current techniques, promising
  directions, and open issues}.
\newblock {\it \bibinfo{journal}{Journal of Visual Communication and Image
  Representation}\/},  {\it \bibinfo{volume}{10}\/}, \bibinfo{pages}{39--62}.
\bibitem[{Saenko et~al.(2010)Saenko, Kulis, Fritz \&
  Darrell}]{saenko2010adapting}
\bibinfo{author}{Saenko, K.}, \bibinfo{author}{Kulis, B.},
  \bibinfo{author}{Fritz, M.}, \& \bibinfo{author}{Darrell, T.}
  (\bibinfo{year}{2010}).
\newblock \bibinfo{title}{Adapting visual category models to new domains}.
\newblock In {\it \bibinfo{booktitle}{European Conference on Computer
  Vision}\/} (pp. \bibinfo{pages}{213--226}).
\newblock \bibinfo{organization}{Springer}.
\bibitem[{Sugiyama et~al.(2008)Sugiyama, Nakajima, Kashima, Buenau \&
  Kawanabe}]{sugiyama2008direct}
\bibinfo{author}{Sugiyama, M.}, \bibinfo{author}{Nakajima, S.},
  \bibinfo{author}{Kashima, H.}, \bibinfo{author}{Buenau, P.~V.}, \&
  \bibinfo{author}{Kawanabe, M.} (\bibinfo{year}{2008}).
\newblock \bibinfo{title}{Direct importance estimation with model selection and
  its application to covariate shift adaptation}.
\newblock In {\it \bibinfo{booktitle}{Advances in Neural Information Processing
  Systems}\/} (pp. \bibinfo{pages}{1433--1440}).
\bibitem[{Sun et~al.(2016)Sun, Feng \& Saenko}]{sun2016return}
\bibinfo{author}{Sun, B.}, \bibinfo{author}{Feng, J.}, \&
  \bibinfo{author}{Saenko, K.} (\bibinfo{year}{2016}).
\newblock \bibinfo{title}{Return of frustratingly easy domain adaptation}.
\newblock In {\it \bibinfo{booktitle}{AAAI Conference on Artificial
  Intelligence}\/}.
\bibitem[{Sun \& Saenko(2016)}]{sun2016deep}
\bibinfo{author}{Sun, B.}, \& \bibinfo{author}{Saenko, K.}
  (\bibinfo{year}{2016}).
\newblock \bibinfo{title}{Deep coral: Correlation alignment for deep domain
  adaptation}.
\newblock In {\it \bibinfo{booktitle}{Computer Vision--ECCV 2016 Workshops}\/}
  (pp. \bibinfo{pages}{443--450}).
\newblock \bibinfo{organization}{Springer}.
\bibitem[{Tzeng et~al.(2017)Tzeng, Hoffman, Saenko \&
  Darrell}]{tzeng2017adversarial}
\bibinfo{author}{Tzeng, E.}, \bibinfo{author}{Hoffman, J.},
  \bibinfo{author}{Saenko, K.}, \& \bibinfo{author}{Darrell, T.}
  (\bibinfo{year}{2017}).
\newblock \bibinfo{title}{Adversarial discriminative domain adaptation}.
\newblock In {\it \bibinfo{booktitle}{Computer Vision and Pattern Recognition
  (CVPR)}\/} (p.~\bibinfo{pages}{4}).
\newblock volume~\bibinfo{volume}{1}.
\bibitem[{Tzeng et~al.(2014)Tzeng, Hoffman, Zhang, Saenko \&
  Darrell}]{tzeng2014deep}
\bibinfo{author}{Tzeng, E.}, \bibinfo{author}{Hoffman, J.},
  \bibinfo{author}{Zhang, N.}, \bibinfo{author}{Saenko, K.}, \&
  \bibinfo{author}{Darrell, T.} (\bibinfo{year}{2014}).
\newblock \bibinfo{title}{Deep domain confusion: Maximizing for domain
  invariance}.
\newblock {\it \bibinfo{journal}{arXiv preprint arXiv:1412.3474}\/}, .
\bibitem[{Zellinger et~al.(2017)Zellinger, Grubinger, Lughofer, Natschl{\"a}ger
  \& Saminger-Platz}]{zellinger2017central}
\bibinfo{author}{Zellinger, W.}, \bibinfo{author}{Grubinger, T.},
  \bibinfo{author}{Lughofer, E.}, \bibinfo{author}{Natschl{\"a}ger, T.}, \&
  \bibinfo{author}{Saminger-Platz, S.} (\bibinfo{year}{2017}).
\newblock \bibinfo{title}{Central moment discrepancy (cmd) for domain-invariant
  representation learning}.
\newblock In {\it \bibinfo{booktitle}{International Conference on Learning
  Representations}\/}.
\bibitem[{Zellinger et~al.(2016)Zellinger, Moser, Chouikhi, Seitner, Nezveda \&
  Gelautz}]{zellinger2016linear}
\bibinfo{author}{Zellinger, W.}, \bibinfo{author}{Moser, B.},
  \bibinfo{author}{Chouikhi, A.}, \bibinfo{author}{Seitner, F.},
  \bibinfo{author}{Nezveda, M.}, \& \bibinfo{author}{Gelautz, M.}
  (\bibinfo{year}{2016}).
\newblock \bibinfo{title}{Linear optimization approach for depth range adaption
  of stereoscopic videos}.
\newblock In {\it \bibinfo{booktitle}{Stereoscopic Displays and Applications
  XXVII}\/}.
\newblock \bibinfo{publisher}{IS$\&$T Electronic Imaging}.
\bibitem[{Zhong et~al.(2010)Zhong, Fan, Yang, Verscheure \&
  Ren}]{zhong2010cross}
\bibinfo{author}{Zhong, E.}, \bibinfo{author}{Fan, W.}, \bibinfo{author}{Yang,
  Q.}, \bibinfo{author}{Verscheure, O.}, \& \bibinfo{author}{Ren, J.}
  (\bibinfo{year}{2010}).
\newblock \bibinfo{title}{Cross validation framework to choose amongst models
  and datasets for transfer learning}.
\newblock In {\it \bibinfo{booktitle}{Joint European Conference on Machine
  Learning and Knowledge Discovery in Databases}\/} (pp.
  \bibinfo{pages}{547--562}).
\newblock \bibinfo{organization}{Springer}.
\bibitem[{Zolotarev(1983)}]{zolotarev1983probability}
\bibinfo{author}{Zolotarev, V.~M.} (\bibinfo{year}{1983}).
\newblock \bibinfo{title}{Probability metrics}.
\newblock {\it \bibinfo{journal}{Teoriya Veroyatnostei i ee Primeneniya}\/},
  {\it \bibinfo{volume}{28}\/}, \bibinfo{pages}{264--287}.

\end{thebibliography}
	
\end{document}